\pdfoutput=1
\documentclass{article}
\usepackage{iclr2027_conference,times}
\iclrfinalcopy

\usepackage{amsmath,amsfonts,bm}

\def\eqref#1{equation~\ref{#1}}

\def\1{\bm{1}}

\DeclareMathAlphabet{\mathsfit}{\encodingdefault}{\sfdefault}{m}{sl}
\SetMathAlphabet{\mathsfit}{bold}{\encodingdefault}{\sfdefault}{bx}{n}

\usepackage{hyperref}
\hypersetup{hidelinks,pdftitle={LoRA's Second Descent Extends Beyond Parameter Parity},pdfauthor={Yueran Ma}}
\usepackage{url}
\usepackage{graphicx}
\usepackage{booktabs}
\usepackage{amsmath,amssymb}
\usepackage{placeins}

\graphicspath{{figures/}{./}}


\newcommand{\rcap}{r_{\mathrm{cap}}}
\newcommand{\rinterp}{r_{\mathrm{interp}}}
\newcommand{\rpeak}{r_{\mathrm{peak}}}
\newcommand{\rvalley}{r_{\mathrm{valley}}}
\newcommand{\rdense}{r_{\mathrm{expr}}}
\newcommand{\rpar}{r_{\mathrm{parity}}}
\newcommand{\srank}{\mathrm{srank}}
\title{LoRA's Second Descent Extends\\ Beyond Parameter Parity}

\author{Yueran Ma\\
The University of Queensland\\
Brisbane, Australia\\
\texttt{s4931778@student.uq.edu.au}}

\begin{document}

\maketitle
\lhead{Preprint. Under review.}

\begin{abstract}
Double descent has sparked considerable interest, with recent work relating it to the data, the model and the learning configuration. Practical fine-tuning commonly involves training a small adapter on top of frozen pretrained weights, as in low-rank adaptation (LoRA). The adapter's rank is the hyperparameter that sets its capacity, yet how this rank relates to double descent has not been well explored. We quantify this relation under label noise on four vision backbones and a $7$B language model with a module-matched rank sweep (MMRS), which extends past full rank and compares every rank with dense fine-tuning of the same modules, paired by seed. On DeiT-Tiny, risk is lowest at rank one and rises sharply as the adapter becomes able to fit the noisy labels, forming an interpolation cliff. Past the peak, risk falls again, but every tested post-peak rank that still saves parameters remains above dense risk. Rank-one LoRA outperforms dense fine-tuning on three of the four vision backbones, consistent with strong regularization at small rank. LoRA thus exhibits a second descent, but matches dense risk only after losing its parameter advantage, first on DeiT-Tiny at four times dense's projection weights. Code is available at \url{https://anonymous.4open.science/r/lora-second-descent-C7B5}.

\end{abstract}

\section{Introduction}
\label{sec:intro}

A practitioner fine-tuning a pretrained backbone on a few thousand possibly mislabeled examples has a capacity dial to set, the rank $r$ of low-rank adaptation (LoRA) \citep{hu2022lora}, and double descent tells them which way to turn it. As capacity grows, test risk follows the classical U-shape, peaks near the interpolation threshold where the model first fits the training data exactly, and then descends again, so that the largest interpolating models often generalize best \citep{belkin2019reconciling,nakkiran2020deep,spigler2019jamming}. Label noise, the very condition that makes rank selection worth worrying about, amplifies the peak \citep{nakkiran2020deep}. Applied to the rank axis, the advice is counterintuitive and actionable: once past the peak, raise the rank.

Whether the rank axis obeys that advice has not been measured. Prior work documents task-dependent rank optima, degradation at high rank, reduced memorization under low-rank updates, and theoretical analyses of rank \citep{jung2025gralora,rathore2025howmuch,steele2026lora,park2026rank,zindari2026lora,bossy2026memorization}, and double descent has been analyzed in linear and deep transfer learning \citep{dar2022double,dar2022frozen}. None combines an operational interpolation threshold on the rank axis, risk measurements past the expressivity limit $r{=}d$ (the model width, at which the adapter can first express any dense update), and a comparison of the tail with the same modules trained densely. The sweep range and the dense reference decide the answer.

\begin{figure}[t]
\centering
\includegraphics[width=0.92\linewidth]{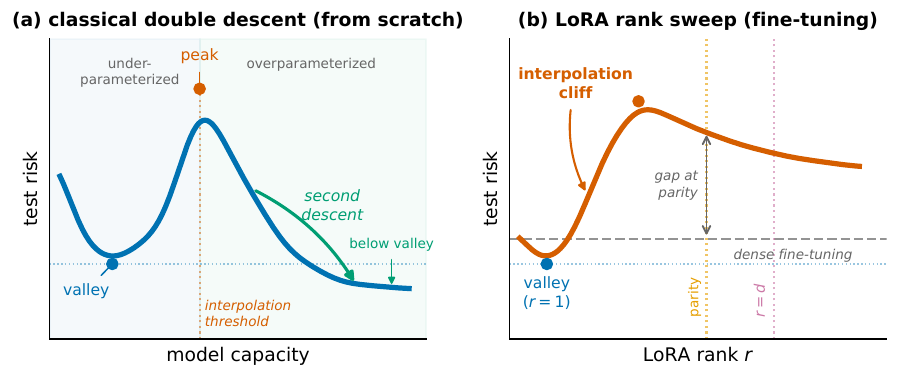}
\caption{\textbf{Two shapes, and why the reference point decides between them.} (a)~On a from-scratch capacity axis, risk peaks at interpolation and then descends past its own pre-peak valley, the only reference such an axis has. (b)~On an adapter's rank axis a second reference exists, dense fine-tuning of the same modules. Parity, the rank at which the adapter has as many trainable weights as dense fine-tuning, arrives before $r{=}d$, the smallest rank that can express every dense update, so the adapter stops saving parameters before it is as expressive as dense. Schematic; measurements in Figure~\ref{fig:main}.}
\label{fig:schematic}
\end{figure}

\textbf{The shape, and where sweeps stop.} Sweeping rank on an ImageNet-pretrained DeiT-Tiny under label noise, we find test risk minimized at $r{=}1$, climbing steeply exactly as ranks acquire the ability to fit the noisy labels, and peaking well below full rank. We call this shape the \textbf{interpolation cliff} (Figure~\ref{fig:schematic}). What matters for measuring the axis is where sweeps conventionally end. Stopped at full rank, the axis has undone $0.033$ of the $0.531$ nats the peak cost and reads as a flat plateau. When the sweep extends to $4d$, risk keeps descending. Whether a sweep sees the second descent depends on how far it is allowed to run (Section~\ref{sec:cliff}).

\textbf{A protocol that sets the reference point.} ``How much of the peak was undone'' has no answer without a reference, and this axis offers two that disagree. Grading the tail against the sweep's own valley is what the double-descent literature does, because a from-scratch width sweep has no unrestricted counterpart to compare against. An adapter axis has one. The \textbf{module-matched rank sweep} (MMRS) runs the dial past full rank and grades every point against dense fine-tuning of the identical modules, sharing seeds, subsets and noise draws, with the trainable-parameter ratio reported beside every risk difference (Section~\ref{sec:method}). On this family the two references have room to disagree, because the window in which the adapter saves parameters closes at parity, $r{=}0.75d$, before it can express every dense update at $r{=}d$ (Figure~\ref{fig:schematic}b).

\textbf{What the tail costs.} Under MMRS the descent is real and its price is legible. Every measured post-peak rank that still saves parameters has higher test negative log-likelihood (NLL) than dense, in every seed at a shared learning rate, so the descent becomes competitive with dense only after the adapter has stopped being cheaper than what it replaces. DeiT-Tiny first meets an equivalence margin of $0.05$ nats at $r{=}3d$, four times dense's projection-weight count, and that result rests on one shared learning rate. The cheap end is a different matter: $r{=}1$ leads dense by $0.40$ and $0.27$ nats on DeiT-Tiny and ViT-Tiny using $0.007\times$ dense's projection-weight count, by far the largest risk reduction per added parameter. Qwen2.5-7B remains above dense at every measured rank while falling below its own pre-peak valley, so the two references answer the same sweep differently (Section~\ref{sec:recovery}). Since full rank supplies the expressivity, the remaining gap is not a function-class deficit; freezing one factor and controlling its spectrum isolates a condition-number effect on risk that disappears when the optimizer operates in update space (Section~\ref{sec:nominal}).

\begin{figure}[t]
\centering
\includegraphics[width=\linewidth]{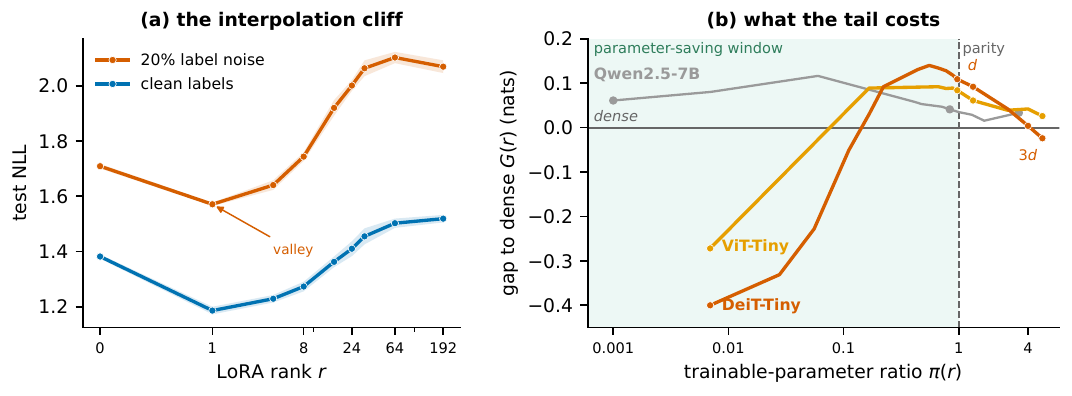}
\caption{\textbf{The interpolation cliff and what it costs to leave.} (a)~Test NLL vs.\ LoRA rank for DeiT-Tiny on CIFAR-100 ($5$k subset), mean$\pm$std over $5$ seeds, $40$ epochs. (b)~DeiT-Tiny, ViT-Tiny and Qwen2.5-7B under the shared learning rate, as paths in the plane of the trainable-parameter ratio $\pi(r)$ and the paired gap $G(r)$ to dense fine-tuning of the same query-key-value (qkv) projections (both defined in Section~\ref{sec:method}). Dots mark $r{=}1$, the last sampled rank below parity, $r{=}d$ and the largest rank; $3d$ marks where DeiT-Tiny first passes the equivalence test of Section~\ref{sec:recovery}.}
\label{fig:main}
\end{figure}

\section{Related work}
\label{sec:related}

\textbf{Double descent.} The risk peak at interpolation and its subsequent descent were established by \citet{belkin2019reconciling}, \citet{spigler2019jamming} and \citet{advani2020highdim}, extended to deep networks, epochs and data by \citet{nakkiran2020deep}, and analyzed precisely in linear and random-features models \citep{hastie2022surprises,mei2022generalization,bartlett2020benign}, as reviewed by \citet{belkin2021fit}. Label noise amplifies the peak \citep{nakkiran2020deep}, regularization can suppress it \citep{nakkiran2021optimal}, and related structure includes triple descent \citep{dascoli2020triple} and grokking \citep{power2022grokking,davies2023unifying}. Nearest to our phenomenon, \citet{xue2024width} report a ``final ascent'' in the widest models once the noise-to-sample ratio is large, a regime our from-scratch transformer reproduces and our ResNet controls do not (Section~\ref{sec:diagnostics}). Closest in spirit to our diagnosis, \citet{curth2023uturn} argue that double descent hinges on what the capacity axis measures and \citet{schaeffer2023double} isolate the ingredients producing the peak. In transfer learning, \citet{dar2022double} analyze linear transfer and \citet{dar2022frozen} sweep frozen-layer count, neither along adapter rank.

\textbf{LoRA and parameter-efficient fine-tuning (PEFT).} LoRA \citep{hu2022lora} builds on intrinsic-dimension measurements of objective landscapes \citep{li2018intrinsic,aghajanyan2021intrinsic}; its rank expressivity is characterized by \citet{zeng2024expressive} and its scaling conventions by \citet{kalajdzievski2023rslora}. That large ranks can hurt is established: accuracy stagnates past ranks of $32$--$64$ \citep{jung2025gralora}, optima are task-dependent \citep{rathore2025howmuch}, LoRA ``learns less and forgets less'' \citep{biderman2024lora}, and low-rank updates memorize less in large language models \citep{bossy2026memorization}; in noisy-label classification, \citet{kim2024robust} find that the capacity limit of PEFT also impedes learning clean examples. Under label noise, theory places the optimal rank lower as noise grows \citep{steele2026lora} and accuracy degrades with rank when noise is strong \citep{zindari2026lora}, while \citet{park2026rank} and \citet{rushka2026rank} analyze rank sufficiency and approximation error. Empirically, \citet{dennis2026procedural} find ranks $16$ to $128$ all trailing full fine-tuning and \citet{albert2025randlora} ask whether the shortfall is parameters or rank, neither under label noise. These establish the two endpoints. What remains unmeasured is the shape between them, where risk peaks relative to the rank at which optimization interpolates, and whether the post-peak branch recovers once the adapter can express any dense update.

\textbf{Effective capacity and optimization geometry.} That LoRA's update carries a stable rank well below its nominal rank is likewise established, as a pathology to engineer away \citep{lion2025polar} and as a sweep locating where validation accuracy turns \citep{awofeso2025stablerank}; we add its placement relative to interpolation and an intervention on it. Parameterization matters beyond expressivity \citep{shuttleworth2025illusion}, and learning rate interacts with rank through initialization and scaling \citep{chen2026lrscaling}, which is why the shared-rate sweep is checked against per-arm rates (each at its own rate) on every backbone (Section~\ref{sec:recovery}, Appendices~\ref{app:envelope} and~\ref{app:e27}). The optimization geometry of the factored parameterization is mapped: freezing one factor \citep{zhang2023lorafa,zhu2024asymmetry} or both \citep{kopiczko2024vera}, preconditioning the gradient \citep{zhang2024riemannian,wang2024lorapro,ghosh2026polora}, and showing that different factorizations of one $\Delta W$ converge differently \citep{castin2026balanced,oh2026sdslora}. Those results concern training and generalization of the factorization; ours concern where the post-peak branch reaches dense risk (Section~\ref{sec:recovery}). \citet{chang2025loramgpo} use ``double descent'' for a transient training-loss divergence, a different object.

\textbf{Memorization and calibration under noise.} Interpolating noisy labels forces memorization \citep{zhang2017rethinking,feldman2020memorization}, and calibration itself depends on overparameterization in high-dimensional models \citep{clarte2023uncertainty}, which is why we separate the risk we measure from the temperature it is read at (Section~\ref{sec:diagnostics}).

\section{The interpolation cliff}
\label{sec:cliff}

Unless noted, this section uses an ImageNet-pretrained DeiT-Tiny on $5$k-example CIFAR-100 subsets with $20\%$ symmetric label noise and LoRA on the fused query-key-value (qkv) projections of the last six blocks (full protocol in Section~\ref{sec:setup}).

\textbf{Risk rises where the adapter begins to fit noise.} Test NLL is minimized at $r{=}1$, below that of the linear probe, so a single rank of adaptation already beats frozen features. The curve then climbs steeply and peaks at $r{=}64$, $0.53$ nats above the minimum (Figure~\ref{fig:main}a). Train accuracy locates the climb: it crosses $99\%$ between $r{=}16$ and $r{=}24$, so risk rises exactly as ranks acquire the ability to fit the noisy training set within budget (Figure~\ref{fig:thresholds}). With clean labels the shape flattens into a mild monotone rise; label noise amplifies this peak as it does the classical one \citep{nakkiran2020deep}. Where classical double descent would begin descending, the noisy curve instead keeps rising to $r{=}64$, and accuracy tells the same story, still below its rank-one level at $r{=}4d$ (Table~\ref{tab:e1}).

\textbf{The pipeline does produce textbook double descent.} From-scratch ResNet-18 width sweeps, run with the same data loading, noise injection, optimizer family and evaluation, peak in the interior and then carry risk below their own pre-peak valley, by $0.196$ nats on CIFAR-10 and, on the identical CIFAR-100 subsets used for fine-tuning, by $0.658$ nats, in every seed on both datasets (Figure~\ref{fig:scratch}). Dataset, sample size and noise draw are not what slows the second descent.

\textbf{Counting parameters says nothing about either landmark.} Three thresholds that classical linear accounts implicitly align come apart here (Figure~\ref{fig:thresholds}): the smallest rank whose trainable-parameter count reaches $n$, the smallest whose train accuracy reaches $99\%$, and the rank of maximal risk. The first threshold is already passed at $r{=}0$, since even the linear head has more trainable parameters than there are training examples. The second arrives at $r{=}24$ and the third at $r{=}64$. Raw parameter count is uninformative about both landmarks, a sharp PEFT instance of the parameter-counting critique \citep{curth2023uturn}; interpolation here has to be measured, not counted.

\textbf{The conventional sweep range hides the far side.} Stopped at $r{=}d$ the tail reads as a flat plateau; extended to $4d$ the same curve keeps descending (Figure~\ref{fig:main}b). Whether a sweep shows the second descent depends on its range, the first thing a protocol has to pin down. The other is the reference point, and this axis supplies two.

\section{The module-matched rank sweep}
\label{sec:method}

\textbf{Problem statement.} Let a pretrained network carry modules $\mathcal M$ with frozen weights $W_m \in \mathbb{R}^{p\times d}$. A parameter-efficient family assigns to each setting $r$ of a capacity dial a set of admissible updates $\Delta W \in \mathcal U(r) \subseteq \mathbb{R}^{p \times d}$, nested so that $\mathcal U(r) \subseteq \mathcal U(r')$ whenever $r < r'$. Nothing below uses more than nestedness. Let $R(r)$ be expected test risk after training at setting $r$ on $n$ examples whose labels carry noise rate $\eta$, let $R_{\mathrm{dense}}$ be the risk of training the same modules over all of $\mathbb{R}^{p\times d}$, and let $\pi(r)$ be the trainable-parameter count at $r$ divided by the count for dense fine-tuning.

With $\rpeak = \arg\max_r R(r)$ and $\rvalley = \arg\min_{r \le \rpeak} R(r)$, a second descent on this axis is a claim about $R$ for $r > \rpeak$, and two quantities make it precise:
\begin{equation}
\mathrm{Rec}(r) \;=\; \frac{R(\rpeak) - R(r)}{R(\rpeak) - R(\rvalley)},
\qquad\qquad
G(r) \;=\; R(r) - R_{\mathrm{dense}} .
\label{eq:two}
\end{equation}
The first grades the tail against the sweep's own valley, so $\mathrm{Rec}{=}1$ is a return to it and $\mathrm{Rec}{>}1$ an overshoot; the second grades it against the same modules trained without the restriction to $\mathcal U(r)$. On a from-scratch capacity axis, the restriction is the model, so no unrestricted counterpart exists. Only $\mathrm{Rec}$ is available, which is why recoveries are read against a valley in the double-descent literature. An adapter axis supplies both, and they can disagree without contradiction, since a sweep may drive $\mathrm{Rec}$ above one while $G$ remains positive at every setting with $\pi(r) < 1$. A second descent certified by $\mathrm{Rec}$ alone can therefore be one that matches dense only where $\pi(r) \ge 1$.

\textbf{Where the axis changes character.} Two settings organize the tail. The \emph{expressivity limit} $\rdense$ is the smallest $r$ with $\mathcal U(r) = \mathbb{R}^{p\times d}$; below it the dial enlarges the function class, above it only the parameterization grows redundant, so a descent past $\rdense$ is a descent under redundancy rather than classical capacity-driven double descent. The \emph{parity point} $\rpar$ is the smallest $r$ with $\pi(r) \ge 1$, past which the family has stopped being parameter-efficient at all. For LoRA on modules of different shapes, $\rdense = \max_m \min(p_m, d_m)$, since the update class is full only once every module's update class is full, while $\rpar$ is set by the total count.

Nothing forces an order on these two, and for LoRA the order is the unhelpful one. A rank-$r$ adapter on a $p \times d$ projection trains $r(p+d)$ weights against dense's $pd$, so $\rpar = pd/(p+d)$ while $\rdense = \min(p,d)$. A fused qkv projection has $p = 3d$, giving
\begin{equation}
\rpar \;=\; \tfrac{3}{4}d \;\;<\;\; d \;=\; \rdense .
\label{eq:order}
\end{equation}
The parameter-saving window closes before the adapter can express every dense update. Any recovery that arrives after $\rdense$ therefore lies past parity, where an adapter no longer saves parameters. This is why $G$ and $\pi$ have to be read together.

\textbf{Interpolation is located, not counted.} The operational interpolation threshold $\rinterp(T)$ is the smallest $r$ whose mean train accuracy reaches $99\%$ within a training budget $T$. Exceeding $n$ trainable parameters, the counting threshold, does not confer interpolation, because a head linear in frozen features is constrained by the feature geometry, not its parameter count. Classical accounts align both thresholds with $\rpeak$; a protocol on this axis has to report the three separately.

\textbf{Protocol.} MMRS combines five choices. (i) Extend the sweep to a multiple of $\rdense$ rather than stopping at it. (ii) Train dense on exactly $\mathcal M$, sharing seeds, subsets and noise draws with every adapter arm, and report paired per-seed differences. (iii) Report $G(r)$ with $\pi(r)$ beside it and mark $\rpar$ on the axis, so no risk gain is read without its parameter price. (iv) Hold the learning rate fixed across arms for the paired sweep, so a per-seed difference reflects the update class rather than the schedule. Then check that per-arm rates do not reverse the ordering, selecting each arm's rate on some seeds and scoring the pair on others; a shared rate can favor whichever arm it suits \citep{chen2026lrscaling}, and a rate scored on the seeds that selected it flatters the higher-variance arm. (v) Locate $\rinterp$ from train accuracy at the budget in use and $\rpeak$ from test risk, and report both.

\section{Experiments}
\label{sec:experiments}

After the shared setup, we read the sweep from its cheap end (Section~\ref{sec:rankone}) through the tail graded against dense (Section~\ref{sec:recovery}) to what separates the two arms at full rank (Section~\ref{sec:nominal}), and then ask whether the cliff belongs to LoRA and how robust its diagnosis is (Sections~\ref{sec:beyond} and~\ref{sec:diagnostics}).

\subsection{Setup}
\label{sec:setup}

\textbf{Task and backbones.} In the main testbed, we fine-tune an ImageNet-1k-pretrained DeiT-Tiny \citep{touvron2021deit,dosovitskiy2021vit,wightman2019timm} of width $d{=}192$ on class-balanced subsets of CIFAR-100 \citep{krizhevsky2009cifar} with $n{=}5{,}000$ training examples. A fraction $\eta$ of training labels is resampled uniformly among the other classes ($20\%$ unless stated). Evaluation uses the untouched $10$k test set. The primary metric is test NLL, the risk addressed by double-descent theory and a more sensitive measure than $0$--$1$ loss, with accuracy and calibrated variants reported alongside it. Each comparison with dense is reported as $G(r)$ from Equation~\ref{eq:two}, which is negative when the adapter has lower risk, with $\pi(r)$ beside it. Bracketed values are $95\%$ bootstrap intervals over paired seeds and are descriptive when only three seeds are shared.

\textbf{Adapter axis.} We wrap the fused qkv projections of the last six transformer blocks with LoRA, so that $Wx \rightarrow Wx+sBAx$ with $A\in\mathbb{R}^{r\times d}$, $B\in\mathbb{R}^{3d\times r}$, and $B$ initialized to zero. The base layer is frozen. Unless stated otherwise, we use $s{=}1$; Appendix~\ref{app:robust} covers $\alpha/r$ scaling and rsLoRA \citep{kalajdzievski2023rslora}. A freshly initialized linear head is always trained ($19{,}300$ parameters), and $r{=}0$ denotes linear probing. Training uses AdamW \citep{loshchilov2019adamw} at a shared learning rate of $5\times10^{-4}$ for $40$ epochs by default, with $5$ seeds for the main sweep; the remaining settings are in Appendix~\ref{app:details}. Seeds control the subset, noise draw and initialization. Parameter ratios $\pi(r)$ compare trainable projection weights, excluding the common head; the vision dense arm also trains the qkv biases, which full-rank updates can absorb on the frozen LayerNorm outputs (Appendix~\ref{app:updateclass}).

\textbf{From-scratch controls.} We train ResNet-18 \citep{he2016resnet} from scratch with the same pipeline across widths, on noisy CIFAR-10 and on the CIFAR-100 subsets used for fine-tuning, with $3$ seeds each (widths and budgets in Table~\ref{tab:blocks}).

\subsection{Rank one yields the most risk reduction per parameter}
\label{sec:rankone}

Measured as the NLL drop from the linear probe divided by the parameters added, the return of the first rank exceeds that of any other rank on the axis, on DeiT-Tiny, ViT-Tiny \citep{steiner2022augreg} and at $7$B. A single rank of adaptation also leads module-matched dense fine-tuning by $0.400$ nats on DeiT-Tiny, $0.272$ on ViT-Tiny and, at per-arm rates, $0.071$ on Swin-Tiny \citep{liu2021swin}, in every seed, using under one percent of dense's projection-weight count (Table~\ref{tab:recovery}, Appendices~\ref{app:window} and~\ref{app:envelope}). On these three backbones $r{=}1$ dominates dense outright, cheaper and lower-risk at once, and needs no argument about the tail. The domination is consistent with strong regularization at small rank.

On DeiT-Small \citep{touvron2021deit} and at $7$B the picture under the shared rate is a trade-off rather than a domination: dense attains the lowest risk, $r{=}1$ comes within $0.074$ and $0.061$ nats of it on under one percent of its projection weights, and at $7$B every sampled rank from $448$ to $900$ falls between, lower-risk than $r{=}1$ while still cheaper than dense (Appendices~\ref{app:window} and~\ref{app:e27}). These ranks are points on a cost--risk frontier that a practitioner with an intermediate budget may want. What none of them does is reach dense, which is what Section~\ref{sec:recovery} measures.

\subsection{The parameter-saving window closes before the descent arrives}
\label{sec:recovery}

Every measured post-peak rank that still saves parameters carries higher risk than dense fine-tuning of the same projections, on DeiT-Tiny, ViT-Tiny and DeiT-Small, in every seed at a shared learning rate (Figure~\ref{fig:main}b, Table~\ref{tab:recovery}, Appendix~\ref{app:window}). The descent itself is genuine, since the DeiT-Tiny and ViT-Tiny sweeps keep falling past the peak once extended to $4d$; what the paired column adds is its price, a gap that narrows only from above and only slowly, so wherever it closes at all it closes far outside the range where the adapter was still cheaper than dense.

\textbf{Inside the window.} We sample five ranks between the peak and $\rpar$ on DeiT-Tiny and three on ViT-Tiny, five seeds each, since those are the only points where a recovery could still retain a parameter advantage (Appendix~\ref{app:window}). None reaches dense. The paired gap remains between $+0.084$ and $+0.140$ nats, roughly four to seven times its seed-to-seed spread, and every seed agrees in sign on both NLL and accuracy. Giving each arm its own learning rate, selected on three seeds, narrows these gaps without closing them: re-run on ten fresh seeds, DeiT-Tiny's $r{=}128$ trails dense by $+0.062$ nats ($95\%$ $t$-interval $[+0.026,+0.099]$, nine of ten seeds), and ViT-Tiny's trails by $+0.065$ on every selection seed (Appendix~\ref{app:envelope}). We have measured the window in which a recovery could still match dense while saving parameters, and it is empty.

\textbf{Outside the window.} At $\rdense$, where either adapter can already express any dense weight update, both backbones remain behind dense. Under the shared rate, DeiT-Tiny passes a paired equivalence test with two one-sided tests (TOST) against a $0.05$-nat margin, a tenth of the cliff's depth, first at $r{=}3d$ (eight seeds), four times dense's projection-weight count; ViT-Tiny does not pass it there. With each arm at its own best rate, dense gains from a higher rate while LoRA's best at $3d$ is the shared one, so the difference there becomes $+0.031$ nats, positive in every seed but inside the margin (Appendix~\ref{app:envelope}); the equivalence at $3d$ is established only under the shared rate.

\begin{figure}[t]
\centering
\includegraphics[width=\linewidth]{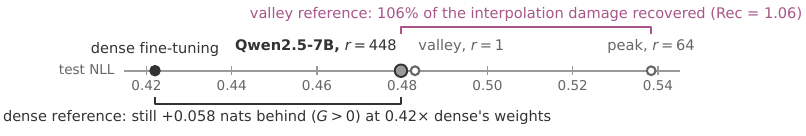}
\caption{\textbf{One rank, two references.} Qwen2.5-7B, three-seed means (Appendix~\ref{app:e27}).}
\label{fig:tworefs}
\end{figure}

\textbf{The same shortfall at $7$B.} To test the finding at language-model scale, we repeat the measurement on Qwen2.5-7B \citep{qwen2025qwen25} over $5{,}000$ AG News examples \citep{zhang2015agnews} with $20\%$ label noise, adapting the qkv projections of the last six blocks at twelve ranks up to $d{=}3584$ against dense fine-tuning of the same modules ($3$ shared seeds; Appendix~\ref{app:e27}). Grouped-query attention makes the projections cheap relative to the adapter and places parity at $r{\approx}1075$, which we sample, so this axis is the clean test of the two references in Equation~\ref{eq:two}. They part company sharply (Figure~\ref{fig:tworefs}). At $r{=}448$ the valley reference reads $\mathrm{Rec}{=}1.06$, a second descent that has already completed, while the dense reference puts the same point $0.058$ nats short, and $G$ remains positive at every measured rank and in every seed, parity included. Read against a valley this sweep recovers fully; read against dense it never arrives.

\begin{table}[t]
\caption{\textbf{Post-peak test NLL and the gap to dense fine-tuning.} We compare LoRA ranks up to $4d$ with dense fine-tuning of the same qkv projections in the same six blocks, under the shared learning rate; the arms share seeds, subsets and noise draws. Params is the trainable projection-weight count relative to dense, so parity falls at $r{=}0.75d$, and $r{=}128$ is one of the post-peak ranks sampled below it (Appendix~\ref{app:window}). Paired diff.\ is the mean per-seed difference, with a $95\%$ bootstrap interval over five or eight shared seeds and a descriptive three-seed range otherwise. Dense is the eight-seed mean; its five- and three-seed means are $1.971$/$1.977$ on DeiT-Tiny and $1.796$/$1.810$ on ViT-Tiny.}
\label{tab:recovery}
\centering
\footnotesize
\setlength{\tabcolsep}{4pt}
\resizebox{\linewidth}{!}{%
\begin{tabular}{lrrcrc}
\toprule
& & \multicolumn{2}{c}{DeiT-Tiny} & \multicolumn{2}{c}{ViT-Tiny} \\
\cmidrule(lr){3-4}\cmidrule(lr){5-6}
adapter rank & params & test NLL & paired diff.\ & test NLL & paired diff.\ \\
\midrule
dense qkv, $k{=}6$ & $1.00\times$ & $1.979$ & --- & $1.802$ & --- \\
\midrule
$r=1$ (valley) & $0.007\times$ & $1.571$ & $-0.400$ $[-.404,-.397]$ & $1.537$ & $-0.272$ $[-.274,-.271]$ \\
$r=64$ / $96$ (peak) & $0.44\times$ / $0.67\times$ & $2.102$ & --- & $1.902$ & --- \\
$r=128$ (below parity) & $0.89\times$ & $2.089$ & $+0.118$ $[+.100,+.134]$ & $1.885$ & $+0.089$ $[+.076,+.102]$ \\
\midrule
$r=d$ (full rank) & $1.33\times$ & $2.069$ & $+0.092$ $[+.083,+.106]$ & $1.871$ & $+0.061$ $[+.047,+.074]$ \\
$r=2d$ & $2.67\times$ & $2.019$ & $+0.042$ $[+.026,+.053]$ & $1.848$ & $+0.039$ $[+.027,+.046]$ \\
$r=3d$ ($8$ seeds) & $4.00\times$ & $1.983$ & $+0.004$ $[-.012,+.019]$ & $1.844$ & $+0.042$ $[+.018,+.066]$ \\
$r=4d$ & $5.33\times$ & $1.953$ & $-0.024$ $[-.038,-.017]$ & $1.836$ & $+0.026$ $[+.014,+.042]$ \\
\bottomrule
\end{tabular}}
\end{table}

\subsection{Optimization coordinates change risk inside a fixed update class}
\label{sec:nominal}

The residual gap at full rank is not a shortage of expressivity. Since $\mathcal U(r) = \mathbb{R}^{p\times d}$ for $r \ge \rdense$, the adapter and dense search the same set from $\rdense$ upward, so any risk difference there stems from how the search is parameterized rather than from what it can reach. The interventions below hold the update class fixed and vary the coordinates, showing that coordinates alone can move risk by roughly the size of the gap. Which property of the coordinates produces the gap in ordinary LoRA, where both factors train jointly, remains open.

\begin{figure}[t]
\centering
\includegraphics[width=\linewidth]{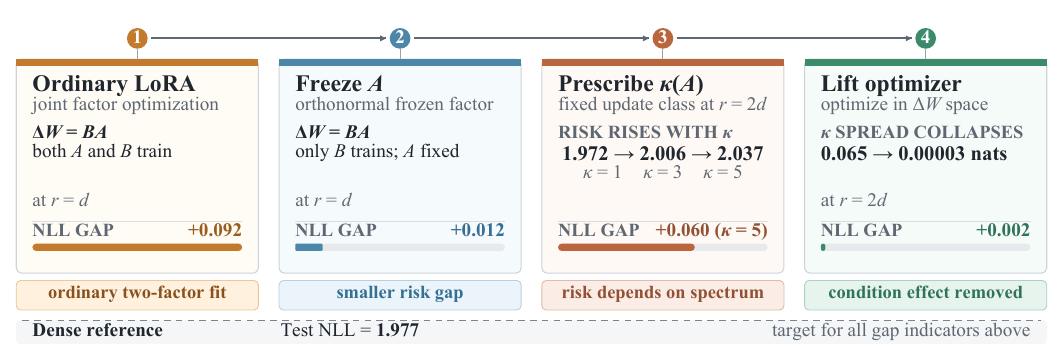}
\caption{\textbf{Risk changes with optimization coordinates.} We compare ordinary LoRA and three interventions on DeiT-Tiny over the same three seeds. The ViT-Tiny results are in Appendix~\ref{app:mech}.}

\label{fig:mechanism}
\end{figure}

\textbf{Different solutions at full rank.} The learned updates differ even where the two procedures search the same set: LoRA's update norm grows with rank while its stable rank, $\|\Delta W\|_F^2/\|\Delta W\|_2^2$, lags far behind nominal rank, and dense reaches a higher stable rank with a smaller norm, a broader spectrum (Appendix~\ref{app:updateclass}). Reaching the same function class does not mean reaching the same solution.

\textbf{Freezing one factor changes the tail.} Holding $A$ fixed makes $B\mapsto BA$ linear while preserving the full weight-update class at $r\geq d$. Frozen-factor methods already use this construction \citep{zhang2023lorafa,zhu2024asymmetry}; here we choose the spectrum of $A$ to control the coordinates. With an orthonormal draw the adapter ends $0.012$ nats above dense at $r{=}d$ and $0.005$ below it at $2d$, on the same three seeds (Figure~\ref{fig:mechanism}). Only $B$ is trained, with $3dr$ projection weights, so its weight count matches dense's at $r{=}d$ and is twice dense's at $2d$.

At $r{=}d$ the orthonormal map is an isometry; for $r>d$, $A^\top A=I$ and the map remains surjective, with redundant directions in $B$. Prescribing the condition number $\kappa(A)\in\{1,3,5\}$ at $r{=}2d$ then keeps the update class fixed while raising risk monotonically in $\kappa$, with stable rank falling in step, on both backbones (Figure~\ref{fig:mechanism}, Table~\ref{tab:frozen_backbones}). Risk depends on the spectrum of a factor that cannot change what the adapter can express. Below parity, freezing $A$ also improves on ordinary LoRA at every tested rank above the valley, by $0.025$--$0.072$ nats on DeiT-Tiny and $0.052$ on ViT-Tiny, yet both frozen arms still finish above dense at $r{=}128$, their largest tested rank below parity.

\textbf{Optimizing in update space removes the condition-number effect.} With $A$ frozen, $\nabla_B\mathcal L=\nabla_{\Delta W}\mathcal L\,A^\top$. A gradient-descent step on $B$ therefore multiplies the update-space gradient by $A^\top A$. AdamW further scales each coordinate through its moment estimates. We keep those estimates directly on $\Delta W$ and lift each step back to $B$ through a left inverse of $A$. This construction makes the update-space step independent of the chosen factor spectrum.

The spectrum then stops mattering. The paired NLL difference between $\kappa{=}5$ and $\kappa{=}1$ collapses from $0.065$ nats to under $10^{-4}$ on DeiT-Tiny and from $0.055$ to under $10^{-3}$ on ViT-Tiny, each lifted arm lands within $0.003$ nats of its dense reference, and stable rank rises to dense's (Table~\ref{tab:frozen_backbones}). The effect of factor geometry on gradients and convergence is established \citep{oh2026sdslora,castin2026balanced}; these experiments show its consequence for test risk at a fixed update class.

In ordinary LoRA, $\kappa(A)$ ranges from $2.8$ to $12.2$ across $r\in\{24,\ldots,128\}$. Projecting $A$ to $\kappa(A){=}1$ after each step closes most of the full-rank gap, but a projection that preserves $BA$ gives higher risk at the same condition number (Appendix~\ref{app:factorcons}), so $\kappa(A)$ alone does not decide the gap. Stable rank is similarly incomplete: a scale-invariant penalty lifts it from $11.2$ to $40.3$ while recovering a tenth of the distance to the valley; at matched stable rank, dense remains $0.070$ nats ahead.

\textbf{Test NLL tracks training NLL more closely than rank.} Pooled over the $175$ fixed-setting runs, test NLL correlates with train NLL at Spearman $\rho{=}{-}0.86$ and with nominal rank at only $0.40$, an ordering that persists after adjustment for the training budget, while higher rank lowers risk within most fully interpolated groups (Appendix~\ref{app:fitstatus}). How closely a run fits the noisy labels predicts its risk better than its rank does.

\subsection{Dense axes recover more of the same cliff}
\label{sec:beyond}

To test whether the cliff is specific to LoRA or to one backbone, we sweep two dense capacity axes, a second PEFT family and three further backbones. The cliff is not an artifact of LoRA or of DeiT-Tiny, while both PEFT families recover less of it than the dense axes (Figures~\ref{fig:axes} and~\ref{fig:breadth}b).

\textbf{Dense capacity axes climb the same cliff.} Unfreezing the last $k$ blocks, or training only their qkv projections densely, produces a cliff deeper than LoRA's from the same head-only start. Both then run a real second descent, recovering $0.474$ and $0.395$ nats of what their peaks cost, against $0.033$ nats on the rank axis by $\rdense$. At $k{=}6$ the module-matched arm trains the same projections as LoRA and finishes about $0.10$ nats below the full-rank adapter, a gap of the size that coordinates alone can produce (Section~\ref{sec:nominal}).

\textbf{A second PEFT family has the flattest tail.} Houlsby-style bottleneck adapters \citep{houlsby2019adapters} on the last six blocks, with bottleneck width $m$ as the capacity axis, trace the full shape from a valley at $m{=}1$ through a peak, then dip by $0.016$ nats of a $0.727$-nat cliff even past twice the model width. Restricted-update families, not LoRA specifically, are what flatten the far side.

\textbf{The rank-one valley holds across backbones.} We test three further backbones with data, noise, optimizer and budget held fixed, each with a module-matched dense arm (Appendix~\ref{app:window}), and two things persist across all of them: the pre-peak valley sits at $r{=}1$, and every other sampled rank inside the parameter-saving window finishes above that valley. ViT-Tiny replicates the cliff in full despite sharing neither recipe nor pretraining data, and remains above dense at every sampled post-peak rank (Table~\ref{tab:recovery}). DeiT-Small has a far shallower cliff and, at $2.7\times$ dense's parameters, finishes below its valley at $2d$ in every seed; against dense it still trails at $r{=}192$, below parity, and at $2d$, by $+0.129$ and $+0.032$ nats under the shared rate and by $+0.056$ and $+0.040$ at per-arm rates, in every seed. Swin-Tiny has no cliff. Its risk rises monotonically from $r{=}1$, which already interpolates and is still the valley, so the rise tracks capacity added past the valley rather than the crossing of interpolation, and there the cheap end wins. At the shared rate, dense overfits and finishes above the linear probe; at each arm's own rate, where all three arms stop short of interpolation, $r{=}1$ and $r{=}384$ still beat dense in every seed, by $0.071$ and $0.055$ nats (Appendix~\ref{app:envelope}).

\subsection{Robustness of the diagnosis}
\label{sec:diagnostics}

\textbf{The rise is not (mostly) miscalibration.} Fitting a temperature on a held-out clean split never used for training \citep{guo2017calibration} leaves $83\%$ of the noisy cliff standing ($0.44$ vs.\ $0.53$ nats, $5$ seeds), rescaling leaves the accuracy drop unchanged, and expected calibration error, Brier score and selective-risk area all degrade across the same range \citep{naeini2015ece,brier1950verification,geifman2017selective,geifman2019biasreduced}. With clean labels calibration removes about half the rise instead (Table~\ref{tab:calib}). Under label noise, the cliff is mostly real error rather than overconfidence.

\textbf{Noise and augmentation change only the cliff's depth.} CIFAR-100N human re-annotation labels \citep{wei2022cifarn} replicate the cliff with larger amplitude. Sweeping $\eta \in \{10, 20, 40, 60\}\%$ keeps the valley pinned at $r{=}1$ while the amplitude grows monotonically, and augmentation shrinks the amplitude without changing the shape or adding a second descent (Appendix~\ref{app:robust}).

\textbf{The peak moves with the training budget.} From $40$ to $80$ epochs it shifts from $r{=}64$ to $r{=}32$, and at $20$ epochs risk is monotone in rank because low ranks have not yet interpolated. The peak reflects interpolation status at the current budget \citep{nakkiran2020deep}, so the protocol locates $\rinterp$ per budget. Neither longer training nor $\alpha/r$ scaling returns the tail to the valley (Appendix~\ref{app:thresholds}).

\textbf{Model selection shrinks the cliff without removing it.} Selecting learning rate, weight decay and stopping epoch on a clean validation split reduces the cliff from $0.53$ to $0.16$ nats. The selected models retain the rank ordering on fresh seeds, and every selected checkpoint precedes interpolation (Appendix~\ref{app:e24}), which is the mechanism: selection escapes the cliff by declining to interpolate, not by crossing it. Because per-rank selection includes no dense arm, it bears on the shape of the axis and not on the gap to dense.

\textbf{No tested setting removes the cliff once ranks interpolate.} Corrupting the frozen backbone only deepens the cliff, and its shape persists through changes in data size, learning rate, weight decay, scaling, adapter placement and dataset (Appendices~\ref{app:envelope}, \ref{app:robust} and~\ref{app:prior}; Table~\ref{tab:e10}).

\textbf{Architecture decides too.} A ResNet completes the descent on these subsets; a width-scaled ViT trained from scratch does not (Appendix~\ref{app:e28}), recovering $0.56$ of $3.23$ nats, the final-ascent regime of \citet{xue2024width}. Dense-block fine-tuning of a transformer reaches $\mathrm{Rec}{=}0.82$, so the rank axis's shallow far side is not a transformer property.

\textbf{Scope.} The experiments cover noisy image and text classification with AdamW, synthetic and human label noise, and budgets up to $160$ epochs. The dense reference is module-matched, training the projections LoRA adapts rather than the whole network; the dense-block axis of Section~\ref{sec:beyond} unfreezes up to six full blocks. Test NLL is the primary measurement, and accuracy follows its ordering inside the parameter-saving window but can diverge at larger ranks. Where LoRA crosses dense depends on the protocol and the grid.

\section{Conclusion}
\label{sec:conclusion}

The rank axis does run a second descent, and a sweep that stops at full rank can miss it. Judging what this descent is worth requires a reference point, which the module-matched rank sweep supplies by grading every setting against dense fine-tuning of the identical modules with the trainable-parameter ratio beside it. Across four vision backbones and a $7$B language model, no measured post-peak rank that still saves parameters reaches dense, and the first equivalence result sits at four times dense's projection weights under a single shared learning rate. Full rank supplies the expressivity, so the remaining gap is a matter of coordinates, which the frozen-factor interventions show can move risk by about the size of the gap. LoRA's parameter advantage is strongest at the cheap end and runs out in the tail; rank-one LoRA dominates dense outright on three of the four vision backbones. Rank keeps shaping generalization long after it stops saving parameters.

\subsubsection*{Acknowledgments}
We thank Marcus Gallagher for helpful discussions and feedback on the manuscript.

\subsubsection*{Reproducibility statement}
Code for every experiment, with the data and checkpoint setup and the MMRS readout, is available at \url{https://anonymous.4open.science/r/lora-second-descent-C7B5}. Hyperparameters are listed in Appendix~\ref{app:details}, and per-run results and figure scripts will be released upon acceptance.

\subsubsection*{AI use statement}
In this work, we used generative AI tools for research ideation and execution, namely to help develop and formalize the conceptual framework, to propose and refine hypotheses, to design experiments and give feedback on them, to implement and run experiments, and to interpret results. We also used them to write the code that builds the noisy-label training sets and to translate drafts. We have not used generative AI tools to generate synthetic datasets. The paper states no theorems, so proving mathematical claims and writing proofs are not applicable to this work, and neither is qualitative or thematic data analysis. Additionally, we used generative AI tools to draft sections of the paper, to create figures, to edit the text for readability, to suggest the structure of the paper, to search for, summarize and check related work and references, and to propose keywords. We have reviewed all AI-assisted work, and we take full responsibility for the content of this paper.
\vspace{-2mm}

\bibliography{references}
\bibliographystyle{iclr2027_conference}

\clearpage
\appendix

\section*{Appendix overview}
Appendix~\ref{app:details} gives the experimental details and a registry of every experiment block. Appendices~\ref{app:cliff} to~\ref{app:breadth} follow the order of the main text, from the interpolation cliff of Section~\ref{sec:cliff} (Appendix~\ref{app:cliff}) through the comparisons with dense fine-tuning in Sections~\ref{sec:rankone} and~\ref{sec:recovery} (Appendix~\ref{app:dense}) and the optimization coordinates of Section~\ref{sec:nominal} (Appendix~\ref{app:mech}) to the breadth and robustness checks of Sections~\ref{sec:beyond} and~\ref{sec:diagnostics} (Appendix~\ref{app:breadth}). Appendix~\ref{app:toy} analyzes a warm-start linear model, and Appendix~\ref{app:tables} collects the full numeric tables.

\section{Experimental details}
\label{app:details}

\textbf{Data.} CIFAR-100/CIFAR-10 \citep{krizhevsky2009cifar} images are upsampled bilinearly to $224^2$ and normalized with ImageNet statistics (ViT-Tiny uses the $0.5$ normalization of its checkpoint). Subsets are class-balanced ($n/C$ per class over $C$ classes, selected by seed-controlled RNG). Symmetric label noise resamples a fraction $\eta$ of training labels uniformly among the remaining $C-1$ classes; the noise pattern is seed-controlled and identical across ranks within a seed. No data augmentation is used in the main sweeps (isolating the capacity axis); E23 repeats the noisy rank sweep with random flip, pad-and-crop ($16$~px at $224^2$ scale, one offset shared per batch), and mixup ($\alpha{=}0.2$).

\textbf{Models.} We use DeiT-Tiny ($5.7$M parameters, patch $16$, width $192$) and DeiT-Small \citep{touvron2021deit} from timm \citep{wightman2019timm} with ImageNet-1k pretrained weights; the classification head is replaced by a freshly initialized linear layer. ViT-Tiny and Swin-Tiny are described in Table~\ref{tab:e26}, and Qwen2.5-7B in Appendix~\ref{app:e27}. LoRA wraps the fused qkv linear layers of the last six blocks; $A$ is initialized with Kaiming-uniform values and $B$ with zeros, so training starts at the pretrained function. The scaling schemes are fixed ($s=1$), standard ($s=\alpha/r$, $\alpha=16$), and rsLoRA \citep[$s=\alpha/\sqrt{r}$, $\alpha=16$;][]{kalajdzievski2023rslora}. In E15, LoRA adapts the fused c\_attn projection of the last six blocks of GPT-2 \citep{radford2019gpt2}; training runs for $12$ epochs at lr $5\times10^{-4}$ on $5{,}000$ AG News examples with $20\%$ label noise, and there is no dense arm.

\textbf{Training.} We train with AdamW (lr $5\times10^{-4}$ unless swept, wd $10^{-2}$, cosine annealing to $0$, batch $64$) and bfloat16 autocast for $40$ epochs unless stated. Each run logs train/test NLL and accuracy every $5$ epochs, final per-example test logits, and per-layer stable ranks of the update.

\textbf{Calibration protocol.} Temperature is fit by grid/LBFGS on a class-balanced $5$k split drawn from the training pool, disjoint from the training subset, with true labels (clean); test labels are never used for fitting.

\textbf{Experiment blocks.} Table~\ref{tab:blocks} summarizes every experiment block, its purpose, the swept setting, and where it appears. Unless stated otherwise, blocks use the main testbed of Section~\ref{sec:setup} (DeiT-Tiny, CIFAR-100 $5$k, $20\%$ noise, $40$ epochs).

\begin{table}[ht]
\centering
\caption{Experiment block summary. Seed counts refer to full sweeps of the listed axis. A dash in the last column marks a block whose results the paper does not report.}
\label{tab:blocks}
\footnotesize
\setlength{\tabcolsep}{4pt}
\resizebox{\linewidth}{!}{%
\begin{tabular}{llllc}
\toprule
block & purpose & sweep / setting & seeds & where \\
\midrule
E1  & main rank sweep & $r \in \{0,\dots,192\}$ incl.\ $96,128$, $\eta \in \{0, 20\%\}$ & 5 & \S\ref{sec:cliff} \\
E2  & budget interaction & $r \times$ epochs $\{20, 80\}$ & 3 & \S\ref{sec:diagnostics} \\
E7  & high-rank, long-time tail & $r \in \{64,192,384\}$, epochs $\{80,160\}$ & 5 & \S\ref{sec:diagnostics} \\
E9  & $\alpha/r$ scaling to interpolation & $r \in \{8,32,192\}$, epochs $\{80,160\}$ & 5 & \S\ref{sec:diagnostics} \\
E5  & from-scratch control & ResNet-18 widths $\{2,\dots,32\}$, CIFAR-10 $12.5$k, $120$ ep & 3 & \S\ref{sec:cliff} \\
E8  & from-scratch, identical data & widths $\{2,\dots,16\}$, CIFAR-100 $5$k, $160$ ep & 3 & \S\ref{sec:cliff} \\
E18 & pretrained, same arch and data & stages $\{0,1,2,4\}$, CIFAR-10 $12.5$k, $20$ ep & 3 & --- \\
E4  & prior corruption & $\alpha \in \{0.75, 0.5, 0.0\}$, $r \in \{4,\dots,192\}$ & 3 & App.~\ref{app:prior} \\
E10 & data size & $n \in \{2.5\mathrm{k}, 10\mathrm{k}\}$ & 2 & \S\ref{sec:diagnostics} \\
E11 & dense blocks axis & last-$k$ blocks, $k \in \{0,1,2,4,6\}$ & 5 & \S\ref{sec:beyond} \\
E13 & module-matched dense qkv & last-$k$ qkv, $k \in \{1,2,6\}$ & 5 & \S\ref{sec:beyond} \\
E22 & dense lr sensitivity & $k{=}6$, lr $\{10^{-4}, 2\times10^{-4}\}$ & 2 & --- \\
E12 & DeiT-Small rank sweep & DeiT-Small, $r \in \{1,24,64,192,384,768\}$ ($384 = d$) & 3--5 & \S\ref{sec:beyond} \\
E16 & bottleneck adapters & width $m \in \{1,4,\dots,256,512\}$ & 3 & \S\ref{sec:beyond} \\
E14 & real human noise & CIFAR-100N labels, $r \in \{0,\dots,384\}$ & 3 & \S\ref{sec:diagnostics} \\
E15 & text check & GPT-2 \citep{radford2019gpt2}, AG News $5$k, $12$ ep, $r \leq 768$ & 3 & App.~\ref{app:tables} \\
E27 & $7$B rank axis vs.\ dense & Qwen2.5-7B, AG News $5$k, $20$ ep, $r \leq 3584$ ($=d$), $+$ dense qkv & 3 & \S\ref{sec:recovery} \\
E28 & from-scratch ViT, same subsets & width $w \in \{24,\dots,384\}$, $100$ ep & 3 & App.~\ref{app:e28} \\
E20 & noise-rate sweep & $\eta \in \{10, 40, 60\}\%$ & 3--5 & \S\ref{sec:diagnostics} \\
E23 & augmentation panel & flip, batch-shared pad-crop, mixup $\alpha{=}0.2$ & 2 & \S\ref{sec:diagnostics} \\
E19 & weight-decay ablation & wd $\in \{0, 0.1\}$ & 2 & App.~\ref{app:envelope} \\
E24 & per-rank hyperparameter envelope & $r \in \{1,\dots,384\} \times$ lr$\times$wd, $80$ ep, val-selected & 1+3 & \S\ref{sec:diagnostics} \\
E26 & backbone external validity & ViT-Tiny, Swin-Tiny; $r \in \{0,1,d/8,d/2,d\}$, ViT-Tiny also $2d/3$, $2d$ & 3 & \S\ref{sec:beyond} \\
E21 & calibration seeds & $r \in \{1,64,192\} \times \eta \in \{0,20\%\}$, seeds $1$--$4$ & 4 & \S\ref{sec:diagnostics} \\
E30 & update norm along the axis & $\|\Delta W\|_F$ at $r \in \{1,\dots,384\}$ and dense qkv $k{=}6$ & 3 (dense 8) & \S\ref{sec:nominal} \\
E29 & stable-rank intervention & $\lambda \in \{0.03,0.3,1.0\}$ at $r \in \{64,192\}$ & 3 & \S\ref{sec:nominal} \\
E31 & frozen orthonormal factor & train $B$ only, $r \leq 2d$, DeiT-Tiny and ViT-Tiny & 3--5 & \S\ref{sec:nominal} \\
E32 & frozen-factor condition number & $\kappa(A) \in \{1,3,5\}$ at $r{=}2d$ & 3--5 & \S\ref{sec:nominal} \\
E33 & rank tail past full rank & $r \in \{3d,4d\}$ on DeiT-Tiny & 3--8 & \S\ref{sec:recovery} \\
E34 & lifted optimizer & AdamW moments on $\Delta W$, $\kappa \in \{1,5\}$ at $r{=}2d$ & 3 & \S\ref{sec:nominal} \\
E35 & ViT-Tiny tail and dense arm & $r \in \{3d,4d\}$ and dense qkv $k{=}6$ & 3--8 & \S\ref{sec:recovery} \\
E36 & $\kappa(A)$ reached by ordinary LoRA & $r \in \{24,\dots,2d\}$, no intervention & 3 & \S\ref{sec:nominal} \\
E37 & constrained factorization & polar projection of $A$ each step, $r \leq 2d$ & 3 & \S\ref{sec:nominal} \\
E39 & function-preserving control & same, with the stretch absorbed into $B$ & 3 & \S\ref{sec:nominal} \\
E38 & symmetric hyperparameter envelope & lr $\{2.5,5,10\}{\times}10^{-4}$, wd $\{0,10^{-2},10^{-1}\}$ & 3 & App.~\ref{app:envelope} \\
E40 & parameter-saving window densified & $r \in \{80,112,140\}$ DeiT-Tiny, $\{112,128,140\}$ ViT-Tiny & 5 & \S\ref{sec:recovery} \\
E41 & dense arms for two more backbones & dense qkv, $k{=}6$, DeiT-Small and Swin-Tiny & 5 & \S\ref{sec:beyond} \\
E42 & each arm at its own learning rate & rate grids on four backbones; DeiT-Tiny $r{=}128$ pair on $10$ fresh seeds & 3, 10 & App.~\ref{app:envelope} \\
E27$'$ & $7$B window densified & $r \in \{512, 900\}$, same protocol as E27 & 3 & App.~\ref{app:e27} \\
K1--K4 & lr, scaling, placement, dataset & see Appendix~\ref{app:robust} & 2--3 & App.~\ref{app:robust} \\
\bottomrule
\end{tabular}}
\end{table}

\section{The interpolation cliff}
\label{app:cliff}

Train accuracy rather than parameter count locates the cliff, and the peak follows the training budget while the tail remains above the valley (Appendix~\ref{app:thresholds}). Trained from scratch through the same pipeline, a ResNet completes the second descent, whereas a transformer does not (Appendix~\ref{app:e28}).

\subsection{Thresholds, training budgets and the high-rank tail}
\label{app:thresholds}

\begin{figure}[ht]
\centering
\includegraphics[width=0.62\linewidth]{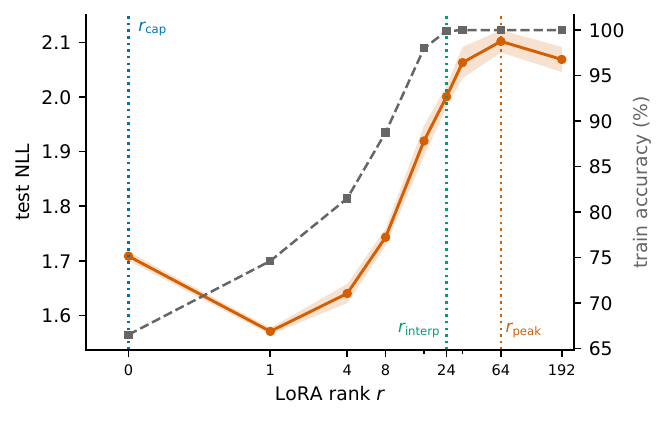}
\caption{\textbf{Three thresholds separate.} Test NLL (orange, left axis) and train accuracy (gray, right axis) vs.\ rank, $20\%$ noise. The counting threshold $\rcap$ is passed at $r{=}0$ (head alone has \mbox{$19{,}300 > n = 5{,}000$} parameters) yet train accuracy is only $67\%$ there; operational interpolation arrives only at $\rinterp=24$; test risk peaks later still at $\rpeak=64$.}
\label{fig:thresholds}
\end{figure}

\textbf{Three thresholds separate.} Figure~\ref{fig:thresholds} places the three thresholds of Section~\ref{sec:cliff} on one axis. The trainable-parameter count exceeds $n$ already at $r{=}0$, train accuracy reaches $99\%$ only at $\rinterp$, and test risk peaks later still at $\rpeak$.

\begin{figure}[ht]
\centering
\includegraphics[width=\linewidth]{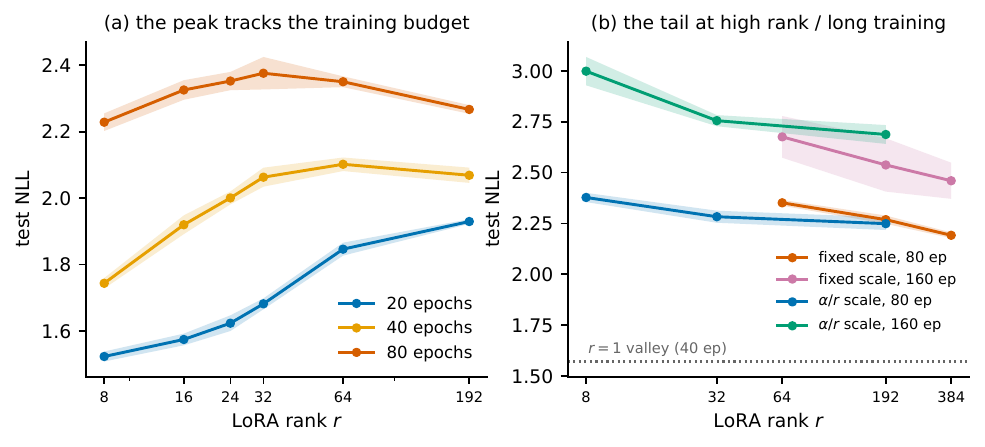}
\caption{\textbf{The tail remains far above the valley.} (a)~Rank sweeps at $20/40/80$ epochs (\mbox{$r\geq8$} shown); the peak tracks whichever ranks are currently reaching interpolation, and the far tail dips but remains high. (b)~High-rank/long-time tail under both scalings (mean$\pm$std, $5$ seeds); mild improvement with rank at fixed budget, degradation with time, and a persistent $\geq 0.62$-nat gap to the $r{=}1$ valley (dotted).}
\label{fig:rankepoch}
\end{figure}

\textbf{The peak follows interpolation.} At $20$ epochs risk rises monotonically across the sampled ranks because the low ranks have not yet interpolated. With longer training the peak moves to lower rank, from $r{=}64$ at $40$ epochs to $r{=}32$ at $80$ (Figure~\ref{fig:rankepoch}a), so the protocol locates it separately at each budget.

\textbf{The tail does not return to the valley.} Under both fixed and $\alpha/r$ scaling, the high-rank tail improves mildly with rank at a fixed budget and degrades with longer training. It remains at least $0.62$ nats above the $r{=}1$ valley (Figure~\ref{fig:rankepoch}b and Table~\ref{tab:tail}).

\subsection{From-scratch controls on the fine-tuning subsets}
\label{app:e28}

\begin{figure}[ht]
\centering
\includegraphics[width=0.55\linewidth]{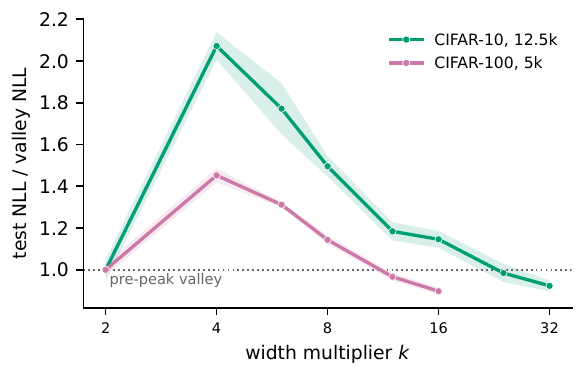}
\caption{\textbf{The same pipeline, swept from scratch, completes the second descent.} ResNet-18 width sweeps on noisy CIFAR-10 and on the identical CIFAR-100 subsets the fine-tuning uses, each divided by its own pre-peak valley risk so the two NLL scales become comparable. Both peak in the interior and finish below the valley, which the DeiT-Tiny rank axis does not do even at $4d$.}
\label{fig:scratch}
\end{figure}

On the noisy fine-tuning subsets the data allow a complete second descent, and the architecture decides whether it appears. ResNet-18 width sweeps through the same pipeline complete the descent on the CIFAR-100 subsets used for fine-tuning and on noisy CIFAR-10 (Figure~\ref{fig:scratch}). A transformer trained from scratch on the same CIFAR-100 subsets reaches $\mathrm{Rec}{=}0.17$, the final-ascent regime of \citet{xue2024width}.

\textbf{Transformer from scratch.} The ViT control of Section~\ref{sec:diagnostics} is a $12$-block transformer trained from scratch without pretrained weights, with embedding width $w$ as the capacity axis and heads scaled as $w/24$, on the same $5{,}000$-example CIFAR-100 subsets and the same $20\%$ noise draw as every fine-tuning axis (lr $10^{-3}$, $100$ epochs, $3$ seeds).

\begin{center}
\footnotesize
\begin{tabular}{rccc}
\toprule
width $w$ & test NLL (mean$\pm$std) & test acc (\%) & train acc (\%) \\
\midrule
$24$  & $4.213 \pm 0.039$ & $12.5$ & $39.5$ \\
$48$  & $7.224 \pm 0.020$ & $9.8$ & $99.7$ \\
$96$  & $7.439 \pm 0.063$ & $10.7$ & $100$ \\
$192$ & $6.809 \pm 0.060$ & $12.3$ & $100$ \\
$288$ & $6.843 \pm 0.044$ & $12.2$ & $100$ \\
$384$ & $6.879 \pm 0.048$ & $12.2$ & $100$ \\
\bottomrule
\end{tabular}
\end{center}

\textbf{The transformer control passes its validity criteria.} Four criteria were fixed before these runs finished, so that a weak control could be discarded rather than reinterpreted, and all four hold. An interior peak exists ($w{=}96$); every post-peak width interpolates (train accuracy $\geq 99\%$); the largest seed standard deviation ($0.063$) is under half the peak-to-valley gap ($3.23$); and post-peak test accuracy does not collapse. Per-seed $\mathrm{Rec}$ lies between $0.15$ and $0.20$. As with the ResNet control, the sweep is read for the shape of the width axis rather than for absolute accuracy.

\FloatBarrier

\section{Comparisons with dense fine-tuning}
\label{app:dense}

Under the shared learning rate, every post-peak rank that still saves parameters trails module-matched dense fine-tuning on the three vision backbones with a cliff (Appendix~\ref{app:window}) and on a $7$B language model (Appendix~\ref{app:e27}). Per-arm learning rates narrow this gap without closing it (Appendix~\ref{app:envelope}), and Appendix~\ref{app:recsens} explains why the tail is graded against dense rather than against the pre-peak valley.

\subsection{The parameter-saving window and the added dense arms}
\label{app:window}

On the three backbones with a cliff, every rank between the peak and parity finishes above module-matched dense in every seed under the shared rate, so no recovery inside the parameter-saving window reaches dense (Table~\ref{tab:window}). Swin-Tiny has no cliff, and its sub-parity ranks finish below a dense arm that overfits at the shared rate (Section~\ref{sec:beyond}).

\textbf{Arms and seeds.} The window between the peak and parity is densified on the two backbones of Table~\ref{tab:recovery} (E40), with five seeds per rank under the shared protocol. Module-matched dense arms, the qkv projections of the last six blocks trained densely as in Section~\ref{sec:setup}, are added with five seeds to DeiT-Small and Swin-Tiny (E41), whose LoRA sweeps are E12 and E26 (Tables~\ref{tab:e12} and~\ref{tab:e26}). Paired differences use the seeds shared by the two arms, dense is the mean over those seeds, and brackets are $95\%$ bootstrap intervals over seeds, descriptive where only three are shared. Parity falls at $r{=}144$ on DeiT-Tiny and ViT-Tiny, $288$ on DeiT-Small and about $430$ on Swin-Tiny, whose adapted blocks have widths $384$ and $768$; the expressivity limit of the set is $768$ there.

\begin{table}[ht]
\caption{\textbf{Gap to module-matched dense fine-tuning under the shared learning rate.} Signs list the per-seed paired differences, $0$ where a pair ties. DeiT-Tiny's peak is the maximum over the main-sweep grid of Table~\ref{tab:e1}; on this denser grid its $r{=}80$ and $96$ lie $0.009$ and $0.003$ nats higher, inside the seed spread (Appendix~\ref{app:recsens}). DeiT-Tiny's $r{=}d$ row pairs the five E1 seeds with dense; Table~\ref{tab:recovery} pairs the three E30 seeds, giving $+0.092$.}
\label{tab:window}
\centering
\footnotesize
\setlength{\tabcolsep}{5pt}
\begin{tabular}{llrccll}
\toprule
backbone & rank & seeds & LoRA NLL & dense NLL & paired diff. & signs \\
\midrule
DeiT-Tiny & $64$ (peak) & 5 & $2.102$ & $1.971$ & $+0.131$ $[+.108,+.149]$ & $+++++$ \\
 & $80$ & 5 & $2.111$ & $1.971$ & $+0.140$ $[+.116,+.161]$ & $+++++$ \\
 & $96$ & 5 & $2.105$ & $1.971$ & $+0.134$ $[+.114,+.150]$ & $+++++$ \\
 & $112$ & 5 & $2.099$ & $1.971$ & $+0.128$ $[+.107,+.151]$ & $+++++$ \\
 & $128$ & 5 & $2.089$ & $1.971$ & $+0.118$ $[+.100,+.134]$ & $+++++$ \\
 & $140$ & 5 & $2.080$ & $1.971$ & $+0.109$ $[+.100,+.123]$ & $+++++$ \\
 & $192$ ($=d$) & 5 & $2.069$ & $1.971$ & $+0.098$ $[+.082,+.117]$ & $+++++$ \\
\midrule
ViT-Tiny & $96$ (peak) & 3 & $1.902$ & $1.810$ & $+0.092$ $[+.055,+.132]$ & $+++$ \\
 & $112$ & 5 & $1.884$ & $1.796$ & $+0.088$ $[+.069,+.105]$ & $+++++$ \\
 & $128$ & 5 & $1.885$ & $1.796$ & $+0.089$ $[+.076,+.102]$ & $+++++$ \\
 & $140$ & 5 & $1.880$ & $1.796$ & $+0.084$ $[+.066,+.098]$ & $+++++$ \\
 & $192$ ($=d$) & 3 & $1.871$ & $1.810$ & $+0.061$ $[+.047,+.074]$ & $+++$ \\
\midrule
DeiT-Small & $1$ (valley) & 5 & $1.396$ & $1.322$ & $+0.074$ $[+.049,+.105]$ & $+++++$ \\
 & $24$ (peak) & 5 & $1.529$ & $1.322$ & $+0.207$ $[+.188,+.226]$ & $+++++$ \\
 & $64$ & 5 & $1.522$ & $1.322$ & $+0.200$ $[+.192,+.213]$ & $+++++$ \\
 & $192$ & 5 & $1.450$ & $1.322$ & $+0.129$ $[+.118,+.143]$ & $+++++$ \\
 & $384$ ($=d$) & 5 & $1.397$ & $1.322$ & $+0.075$ $[+.064,+.083]$ & $+++++$ \\
 & $768$ ($=2d$) & 3 & $1.363$ & $1.331$ & $+0.032$ $[+.025,+.044]$ & $+++$ \\
\midrule
Swin-Tiny & $0$ (probe) & 3 & $1.483$ & $1.530$ & $-0.047$ $[-.055,-.042]$ & $---$ \\
 & $1$ (valley) & 3 & $1.353$ & $1.530$ & $-0.177$ $[-.186,-.165]$ & $---$ \\
 & $96$ & 3 & $1.378$ & $1.530$ & $-0.153$ $[-.186,-.132]$ & $---$ \\
 & $384$ & 3 & $1.512$ & $1.530$ & $-0.018$ $[-.034,+.000]$ & $--0$ \\
 & $768$ ($=d$) & 3 & $1.602$ & $1.530$ & $+0.072$ $[+.050,+.096]$ & $+++$ \\
\bottomrule
\end{tabular}
\end{table}

\subsection{Each arm at its own learning rate}
\label{app:envelope}
Per-arm rates narrow the gap inside the parameter-saving window without closing it, and past parity dense remains ahead of LoRA in every seed through $r{=}3d$.

\textbf{Search.} Each arm's learning rate is searched separately, with three seeds per setting and the same $40$-epoch budget (E38, E42). The grid is refined to half-steps, factors of about $\sqrt{2}$, around dense's optimum and the main sub-parity rank ($r{=}128$ on DeiT-Tiny and ViT-Tiny, $r{=}192$ on DeiT-Small), and is spaced by factors of two elsewhere. The best rate is the one with the lowest mean final test NLL on those three seeds. Table~\ref{tab:envelope} reports each arm at its best rate against dense at its own; Appendix~\ref{app:e24} separately reports models selected by validation NLL and stopping epoch.

\begin{table}[ht]
\caption{\textbf{Each arm at its own learning rate.} Best rate and final test NLL on the three seeds used to select it, and the paired difference from dense at dense's best rate, with per-seed signs. The row marked fresh contains re-runs of the selected DeiT-Tiny pair on ten seeds excluded from selection, with a $95\%$ $t$-interval. Train accuracy is $100\%$ at every best rate except DeiT-Small $r{=}1$ ($77\%$) and the three Swin-Tiny arms (dense $90\%$, $r{=}1$ $72\%$, $r{=}384$ $79\%$).}
\label{tab:envelope}
\centering
\footnotesize
\setlength{\tabcolsep}{5pt}
\begin{tabular}{llrlrl}
\toprule
backbone & arm & params & best rate & NLL & minus best dense \\
\midrule
DeiT-Tiny & dense qkv & $1.00\times$ & $1.0{\times}10^{-3}$ & $1.948$ & --- \\
 & $r{=}96$ & $0.67\times$ & $2.0{\times}10^{-3}$ & $1.968$ & $+0.019$, $++-$ \\
 & $r{=}128$ & $0.89\times$ & $2.0{\times}10^{-3}$ & $1.945$ & $-0.003$, $++-$ \\
 & $r{=}128$, fresh & $0.89\times$ & $2.0{\times}10^{-3}$ & $2.009$ & $+0.062$ $[+.026,+.099]$, 9 of 10 \\
 & $r{=}140$ & $0.97\times$ & $1.4{\times}10^{-3}$ & $1.952$ & $+0.004$, $++-$ \\
 & $r{=}d$ & $1.33\times$ & $1.0{\times}10^{-3}$ & $1.987$ & $+0.039$, $+++$ \\
 & $r{=}3d$ & $4.00\times$ & $5.0{\times}10^{-4}$ & $1.979$ & $+0.031$, $+++$ \\
 & $r{=}4d$ & $5.33\times$ & $5.0{\times}10^{-4}$ & $1.953$ & $+0.004$, $++-$ \\
\midrule
ViT-Tiny & dense qkv & $1.00\times$ & $7.0{\times}10^{-4}$ & $1.786$ & --- \\
 & $r{=}96$ & $0.67\times$ & $1.0{\times}10^{-3}$ & $1.865$ & $+0.079$, $+++$ \\
 & $r{=}128$ & $0.89\times$ & $1.0{\times}10^{-3}$ & $1.851$ & $+0.065$, $+++$ \\
 & $r{=}d$ & $1.33\times$ & $1.0{\times}10^{-3}$ & $1.826$ & $+0.040$, $+++$ \\
\midrule
DeiT-Small & dense qkv & $1.00\times$ & $7.0{\times}10^{-4}$ & $1.323$ & --- \\
 & $r{=}1$ & $0.003\times$ & $2.5{\times}10^{-4}$ & $1.288$ & $-0.034$, $---$ \\
 & $r{=}192$ & $0.67\times$ & $1.4{\times}10^{-3}$ & $1.378$ & $+0.056$, $+++$ \\
 & $r{=}2d$ & $2.67\times$ & $5.0{\times}10^{-4}$ & $1.363$ & $+0.040$, $+++$ \\
\midrule
Swin-Tiny & dense qkv & $1.00\times$ & $5.0{\times}10^{-5}$ & $1.242$ & --- \\
 & $r{=}1$ & $0.002\times$ & $1.0{\times}10^{-4}$ & $1.171$ & $-0.071$, $---$ \\
 & $r{=}384$ & $0.89\times$ & $5.0{\times}10^{-5}$ & $1.187$ & $-0.055$, $---$ \\
\bottomrule
\end{tabular}
\end{table}
\FloatBarrier

\textbf{Inside the window, the tie does not survive fresh seeds.} On DeiT-Tiny the three selection seeds put $r{=}96$, $128$ and $140$ within seed noise of dense. A best rate scored on the seeds that chose it is optimistic, however. At $r{=}96$ and $128$ LoRA's optimum sits one half-step below a rate at which it degrades sharply ($2.8{\times}10^{-3}$), and at $r{=}128$ one full step below divergence ($4{\times}10^{-3}$). On ten fresh seeds, LoRA at $r{=}128$ trails dense in nine when both use their selected rates (Table~\ref{tab:envelope}; paired $t$-test $p{=}0.004$, exact sign-flip $p{=}0.008$), and the difference fails equivalence at the $0.05$-nat margin. On ViT-Tiny and DeiT-Small every post-peak sub-parity comparison remains positive in every seed.

\textbf{Past parity, the arms first draw level at $4d$.} On DeiT-Tiny the gap remains positive in every seed at $r{=}d$ and $3d$, and three seeds do not resolve a difference at $4d$, where LoRA's grid is one step coarser than dense's.

\textbf{The cheap end still wins at per-arm rates.} On Swin-Tiny dense overfits at the shared rate, ten times its own best rate; at each arm's own rate all three arms stop short of interpolation, and the cheap end still wins in every seed. On DeiT-Small, $r{=}1$ at the lowest rate tried also edges past dense on the three selection seeds without interpolating; it is not counted among the dominations of Section~\ref{sec:rankone}, which also hold under the shared protocol.

\textbf{The gap and the cliff persist at every weight-decay coefficient tried.} At each of the coefficients $0$, $10^{-2}$ and $10^{-1}$, LoRA at $r{=}d$ (two seeds) finishes $0.08$ nats above dense (three seeds), and raising the coefficient from $10^{-2}$ to $10^{-1}$ lowers the test NLL of both arms by $0.028$ nats. At coefficients $0$ and $10^{-1}$ the LoRA sweep over $r\in\{1,8,24,64,192\}$ (E19, two seeds) keeps its valley at $r{=}1$ and its peak at $r{=}64$. Dense decays the trained weight $W+\Delta W$, whereas ordinary LoRA decays its factors, so equal coefficients define different regularizers.

\subsection{The rank axis on a 7B language model}
\label{app:e27}
The $7$B rank axis is the clearest case for reading the tail against a module-matched baseline rather than against the pre-peak valley. Every sampled rank, parity included, remains behind dense in all three seeds, while the valley reference reads $\mathrm{Rec}>1$ at every sampled rank from $r{=}448$ on.

\textbf{Setup.} Qwen2.5-7B \citep{qwen2025qwen25}, with $d{=}3584$ and grouped-query attention with key and value width $512$, is fine-tuned for sequence classification on $5{,}000$ AG News examples \citep{zhang2015agnews} with $20\%$ symmetric label noise, sequence length $128$, a freshly initialized linear head, AdamW at lr $6.6\times10^{-5}$, weight decay $10^{-2}$, batch $16$, $20$ epochs, bfloat16. LoRA wraps the query, key and value projections of the last six blocks as three separate modules, so the adapter trains $15{,}360r$ parameters per block against dense's $16{,}515{,}072$ and parity falls at $r{\approx}1075$. The dense arm trains the weights of those projections and freezes their biases, as does LoRA. Both arms train the same classification head with the same optimizer settings. Seeds control the subset, the noise draw, the head init and the adapter init, and are shared across arms, so the differences below are paired. Each run takes about an hour on one A100, and the $r{=}512$ and $r{=}900$ runs take under an hour each on one RTX 5090 under identical settings. In the table, $\srank$ is the stable rank $\|\Delta W\|_F^2/\|\Delta W\|_2^2$ of the update, averaged over adapted modules.

\begin{center}
\footnotesize
\setlength{\tabcolsep}{4pt}
\resizebox{\linewidth}{!}{%
\begin{tabular}{lrccccr}
\toprule
arm & params & test NLL & test acc (\%) & train acc (\%) & $\srank$ & paired diff.\ \\
\midrule
dense qkv, $k{=}6$ & $1.00\times$ & $0.4220 \pm 0.0212$ & $87.23$ & $100$ & $15.0$ & --- \\
\midrule
$r=0$ (head only)  & $0.000\times$ & $0.4938 \pm 0.0114$ & $85.37$ & $80.3$ & --- & $+0.0718$ $[+.064,+.086]$ \\
$r=1$ (valley)     & $0.001\times$ & $0.4830 \pm 0.0112$ & $85.79$ & $80.7$ & $1.0$ & $+0.0610$ $[+.053,+.076]$ \\
$r=8$              & $0.007\times$ & $0.5024 \pm 0.0198$ & $83.32$ & $89.8$ & $1.3$ & $+0.0803$ $[+.069,+.087]$ \\
$r=64$ (peak)      & $0.060\times$ & $0.5384 \pm 0.0197$ & $82.98$ & $100$ & $1.5$ & $+0.1164$ $[+.102,+.126]$ \\
$r=448$            & $0.417\times$ & $0.4797 \pm 0.0108$ & $85.95$ & $100$ & $1.9$ & $+0.0577$ $[+.049,+.073]$ \\
$r=512$ (k/v limit) & $0.476\times$ & $0.4750 \pm 0.0140$ & $86.04$ & $100$ & $2.0$ & $+0.0530$ $[+.045,+.065]$ \\
$r=768$            & $0.714\times$ & $0.4691 \pm 0.0174$ & $86.56$ & $100$ & $2.1$ & $+0.0470$ $[+.035,+.057]$ \\
$r=900$            & $0.837\times$ & $0.4631 \pm 0.0227$ & $86.70$ & $100$ & $2.2$ & $+0.0411$ $[+.031,+.049]$ \\
$r=1075$ (parity)  & $1.000\times$ & $0.4570 \pm 0.0220$ & $87.03$ & $100$ & $2.3$ & $+0.0350$ $[+.029,+.040]$ \\
$r=1400$           & $1.302\times$ & $0.4510 \pm 0.0211$ & $87.34$ & $100$ & $2.4$ & $+0.0289$ $[+.023,+.033]$ \\
$r=1792$           & $1.667\times$ & $0.4376 \pm 0.0226$ & $87.75$ & $100$ & $2.5$ & $+0.0155$ $[+.013,+.019]$ \\
$r=3584~({=}d)$    & $3.333\times$ & $0.4554 \pm 0.0194$ & $87.71$ & $100$ & $3.1$ & $+0.0334$ $[+.030,+.037]$ \\
\bottomrule
\end{tabular}}
\end{center}

Test NLL is mean$\pm$std over the three seeds, params is the trainable projection-weight count relative to dense, and paired diff.\ is the mean per-seed NLL difference from dense; the bracketed range is the spread under paired resampling of the three seeds and is descriptive rather than a calibrated interval. Every entry is positive with all three seeds agreeing in sign.

\textbf{The valley reference would call the descent complete.} Endpoint-normalized recovery (Appendix~\ref{app:recsens}) reads $\mathrm{Rec}(448)=1.06$ and $\mathrm{Rec}(1792)=1.82$ on this axis, which taken alone would report a completed second descent at $0.42\times$ dense's parameters, while the paired column above shows both points short of dense in every seed.

\textbf{Full rank on the key and value projections does not reach dense.} Grouped-query attention makes the $512$-wide key and value projections fully expressive from $r{=}512$, below the parity of the whole set, while the query projection needs all $3584$. Even at $r{=}512$, on under half of dense's projection weights, every seed remains behind dense.

\textbf{The ordering holds at each arm's best tested rate.} Dense, the parity rank and $r{=}1792$, the rank that comes closest to dense, all reach their lowest measured risk at the sweep rate of $6.6{\times}10^{-5}$ (table below). Taking each arm at its best tested rate leaves the gaps of $0.035$ nats at parity and $0.016$ nats at $r{=}1792$ unchanged, and at $r{=}448$ a higher rate gives far higher risk than the sweep rate does.

\begin{center}
\footnotesize
\begin{tabular}{lccccc}
\toprule
& \multicolumn{5}{c}{test NLL at learning rate} \\
\cmidrule(lr){2-6}
arm & $3.3{\times}10^{-5}$ & $6.6{\times}10^{-5}$ & $9.4{\times}10^{-5}$ & $1.3{\times}10^{-4}$ & $1.9{\times}10^{-4}$ \\
\midrule
dense qkv, $k{=}6$ & $0.4943$ & $0.4220$ & --- & $0.4255$ & --- \\
$r{=}448$ & --- & $0.4797$ & --- & --- & $0.6565$ \\
$r{=}1075$ (parity) & $0.4806$ & $0.4570$ & --- & $0.5763$ & --- \\
$r{=}1792$ & $0.4546$ & $0.4376$ & $0.4970$ & --- & --- \\
\bottomrule
\end{tabular}
\end{center}

\subsection{Endpoint-normalized recovery as a secondary diagnostic}
\label{app:recsens}

Conventional double-descent curves are often summarized by how much of the interpolation peak the far branch undoes, which is $\mathrm{Rec}$ of Equation~\ref{eq:two}. That summary is well defined only once a stopping point is named, and we read it at successive endpoints in the table below. On the two rank axes swept to $4d$, $\mathrm{Rec}$ keeps rising well past the point where the adapter can express any dense update, and its value at that point is not final.

\begin{center}
\footnotesize
\begin{tabular}{lccccc}
\toprule
axis & gap (nats) & $\mathrm{Rec}@d$ & $\mathrm{Rec}@2d$ & $\mathrm{Rec}@3d$ & $\mathrm{Rec}@4d$ \\
\midrule
LoRA rank, DeiT-Tiny & $0.531$ & $0.06$ & $0.16$ & $0.22$ & $0.28$ \\
LoRA rank, ViT-Tiny  & $0.365$ & $0.09$ & $0.15$ & $0.16$ & $0.18$ \\
\bottomrule
\end{tabular}
\end{center}

\textbf{Returning to the valley and matching dense are different targets.} The DeiT-Tiny axis reaches neither by $d$, having undone under a tenth of its own damage while sitting $0.092$ nats behind dense. By $3d$ it is within $0.004$ nats of dense while still having undone under a quarter of the peak, so on these data the valley and the dense baseline are far apart, and the second is the one a practitioner can act on.

\begin{table}[ht]
\centering
\caption{The three points behind every $\mathrm{Rec}$ outside Appendix~\ref{app:prior}. Seed-mean test NLL at the valley, the peak and the last point measured, with the grid position in parentheses; gap is peak minus valley. $\mathrm{Rec}@d$ reads a rank axis at its expressivity limit and $\mathrm{Rec}@\mathrm{end}$ at the last point measured; the two coincide where the sweep stops at $d$. $\mathrm{Rec}@\mathrm{prev}$ reads each axis at the grid point before its end. Dense blocks and dense qkv end at $k{=}6$, the whole adapted stack, which is the expressivity endpoint their $\mathrm{Rec}$ shares with $\mathrm{Rec}@d$ on the rank axes. Swin-Tiny (E26) and GPT-2 (E15) rise to the last rank measured and have no far side to read.}
\label{tab:rec}
\footnotesize
\setlength{\tabcolsep}{4pt}
\resizebox{\linewidth}{!}{%
\begin{tabular}{llccccccc}
\toprule
axis & variable & valley & peak & end & gap & $\mathrm{Rec}@d$ & $\mathrm{Rec}@\mathrm{end}$ & $\mathrm{Rec}@\mathrm{prev}$ \\
\midrule
LoRA rank, DeiT-Tiny (E1, E30, E33) & $r$ & $1.571$ ($1$) & $2.102$ ($64$) & $1.953$ ($4d$) & $0.531$ & $0.06$ & $0.28$ & $0.22$ \\
LoRA rank, ViT-Tiny (E26, E35) & $r$ & $1.537$ ($1$) & $1.902$ ($96$) & $1.836$ ($4d$) & $0.365$ & $0.09$ & $0.18$ & $0.16$ \\
LoRA rank, DeiT-Small (E12) & $r$ & $1.396$ ($1$) & $1.529$ ($24$) & $1.363$ ($2d$) & $0.133$ & $0.99$ & $1.25$ & $0.99$ \\
LoRA rank, CIFAR-100N human noise (E14) & $r$ & $1.743$ ($1$) & $2.408$ ($64$) & $2.350$ ($2d$) & $0.664$ & $0.04$ & $0.09$ & $0.04$ \\
LoRA rank, Qwen2.5-7B (E27) & $r$ & $0.483$ ($1$) & $0.538$ ($64$) & $0.455$ ($d$) & $0.055$ & $1.50$ & $1.50$ & $1.82$ \\
\midrule
Dense blocks (E11) & $k$ & $1.709$ ($0$) & $2.288$ ($1$) & $1.814$ ($6$) & $0.579$ & --- & $0.82$ & $0.57$ \\
Dense qkv (E13) & $k$ & $1.709$ ($0$) & $2.366$ ($2$) & $1.971$ ($6$) & $0.657$ & --- & $0.60$ & $0.00$ \\
Bottleneck adapters (E16) & $m$ & $1.611$ ($1$) & $2.339$ ($256$) & $2.323$ ($512$) & $0.727$ & --- & $0.02$ & $0.00$ \\
\midrule
ViT width, from scratch (E28) & $w$ & $4.213$ ($24$) & $7.439$ ($96$) & $6.879$ ($384$) & $3.226$ & --- & $0.17$ & $0.18$ \\
ResNet-18 width, CIFAR-10 $12.5$k (E5) & $k$ & $2.560$ ($2$) & $5.304$ ($4$) & $2.364$ ($32$) & $2.743$ & --- & $1.07$ & $1.01$ \\
ResNet-18 width, CIFAR-100 $5$k (E8) & $k$ & $6.433$ ($2$) & $9.341$ ($4$) & $5.774$ ($16$) & $2.908$ & --- & $1.23$ & $1.07$ \\
\bottomrule
\end{tabular}}
\end{table}

$\mathrm{Rec}$ requires naming three points on a curve, and each choice is open to debate. Table~\ref{tab:rec} records the three for every axis outside Appendix~\ref{app:prior}, and the paragraphs below re-derive $\mathrm{Rec}$ under the alternatives.

\textbf{Pinning a common valley leaves the ranking unchanged.} Each axis is normalized by its own peak-to-valley gap, and the axes do not reach the same valley. LoRA's $r{=}1$ valley lies below the head-only model, which no dense axis beats, so the low-rank axis is scored against a deeper valley. Every DeiT-Tiny axis on the synthetic-noise subsets shares the head-only model at $r{=}0$ or $k{=}0$, and the human-noise and ViT-Tiny axes each have their own. Pinning each valley at its head-only model puts the axes on a common footing. It raises $\mathrm{Rec}@d$ on the DeiT-Tiny, human-noise and ViT-Tiny rank axes by at most $0.02$ and leaves the dense axes unchanged, since their valleys already sit at $k{=}0$. Pinned at its own head-only model, the $7$B axis would read $\mathrm{Rec}@d\approx1.86$ instead of $1.50$, which widens the disagreement between its two references (Appendix~\ref{app:e27}). Axes whose sweeps have no $r{=}0$ point (DeiT-Small, bottleneck adapters and the from-scratch controls) cannot be re-pinned and are left as measured.

\textbf{Another choice of peak moves $\mathrm{Rec}$ by at most $0.02$.} Where two grid points sit within a standard deviation of each other, which one is called the peak is arbitrary. This happens on three axes, namely ViT-Tiny ($r{=}24$ and $96$), DeiT-Small ($r{=}24$ and $64$), and the main LoRA axis, where the densified window of Appendix~\ref{app:window} places $r{=}80$ and $96$ within the seed spread of $r{=}64$. On both dense axes and the bottleneck-adapter axis, no other grid point comes within a standard deviation of the peak.

\textbf{Shortening the tail lowers most recoveries.} $\mathrm{Rec}$ is read at the last point measured. To see how much rests on that point, the last column of Table~\ref{tab:rec} gives $\mathrm{Rec}$ one grid point earlier on every axis. Apart from the $7$B axis and the from-scratch ViT, whose tails have already turned back up, every axis recovers less at that earlier point. The ResNet controls still exceed $1$, whereas the bottleneck-adapter and dense-qkv axes fall to zero because each has only one point past its peak. Because the recovery an axis reports depends on how far its sweep runs, Table~\ref{tab:rec} records where every sweep ends, and $\mathrm{Rec}$ is read as final only for axes that reach their expressivity limit.

\FloatBarrier

\section{Optimization coordinates}
\label{app:mech}

The interventions of Section~\ref{sec:nominal} hold the update class fixed and change only the coordinates. Appendix~\ref{app:updateclass} establishes that full-rank LoRA and dense share one update class and counts their parameters, and Appendix~\ref{app:frozenopt} shows that the factor spectrum moves risk under ordinary AdamW but not under update-space steps. Appendix~\ref{app:factorcons} finds that neither the condition number of ordinary LoRA's factor nor the stable rank of its update decides the gap to dense by itself, and Appendix~\ref{app:fitstatus} shows that training NLL orders test risk more closely than nominal rank does.

\subsection{Update class and parameter counts}
\label{app:updateclass}

The vision dense arm trains each qkv layer's weight and bias. LoRA freezes the original layer and adds $BAx$. Both train the classification head. For six DeiT-Tiny or ViT-Tiny blocks, dense therefore trains $686{,}308$ parameters including the head, of which $3{,}456$ are qkv biases. The projection-weight ratios used in the main text compare $4dr$ with $3d^2$ per block. Including the dense biases moves parity from $r{=}144$ to $144.75$, leaving the sampled ranks on the same side of the boundary.

Full-rank LoRA can also express the dense bias changes because the qkv input is a frozen LayerNorm output. Write
\[
z=\gamma\odot\hat x+\beta,\qquad \mathbf{1}^{\top}\hat x=0,
\qquad c=\sum_{i=1}^{d}\frac{\beta_i}{\gamma_i}.
\]
If every $\gamma_i\ne0$ and $c\ne0$, the vector $u_i=(c\gamma_i)^{-1}$ satisfies $u^\top z=1$. A weight and bias update can then be written as a weight update alone,
\[
\Delta Wz+\delta b=(\Delta W+\delta b\,u^\top)z.
\]
Every adapted LayerNorm in the two checkpoints satisfies these conditions. The smallest $|c|$ is $0.0301$ on DeiT-Tiny and $0.0589$ on ViT-Tiny. At $r\geq d$, the matrix on the right is representable by $BA$. The function classes coincide on these normalized inputs, although the corresponding parameter norms and optimization paths can differ. On Qwen2.5-7B, both arms freeze the projection biases.

\textbf{At full rank LoRA's update ends larger but spectrally narrower than dense's.} The table below averages the six adapted projections over the three seeds of the update-norm runs (E30). At $r{=}d$ LoRA's update is $1.3$ times as large as dense's in Frobenius norm but has under two fifths of its stable rank, and at $2d$ its stable rank is still half of dense's. Along the main sweep the stable rank of LoRA's update grows sublinearly with nominal rank (Figure~\ref{fig:unifying}, left).

\begin{center}
\footnotesize
\begin{tabular}{lcccc}
\toprule
& LoRA $r{=}1$ & LoRA $r{=}d$ & LoRA $r{=}2d$ & dense qkv \\
\midrule
$\|\Delta W\|_F$ & $2.24$ & $7.33$ & $8.43$ & $5.64$ \\
stable rank & $1.0$ & $11.2$ & $14.6$ & $29.2$ \\
\bottomrule
\end{tabular}
\end{center}

\subsection{Frozen factors and update-space optimization}
\label{app:frozenopt}

Under ordinary AdamW the spectrum of a frozen factor moves risk at a fixed update class, and update-space steps remove this effect on both backbones.

\textbf{Frozen factor (E31).} We draw $A\in\mathbb{R}^{r\times d}$ using \texttt{nn.init.orthogonal\_} and freeze it. Only $B\in\mathbb{R}^{3d\times r}$ is trained, starting at zero. This uses $3dr$ projection weights, equal to dense at $r{=}d$ and twice dense at $2d$. Including the head, the six-block counts are $682{,}852$ and $1{,}346{,}404$, respectively. The same counts apply when a custom optimizer updates $B$.

For square orthogonal $A$, the map $B\mapsto BA$ preserves the Frobenius norm. When $r>d$, $A^\top A=I$ and the map is surjective but has a nontrivial kernel. It is an isometry on matrices of the form $B=CA^\top$, rather than on the whole parameter space.

\textbf{Prescribed condition number (E32).} Take the thin SVD $A=U\Sigma V^\top$ of the orthonormal draw, replace its singular values by $\mathrm{logspace}(0,-\log_{10}\kappa,k)$ for $k=\min(r,d)$, and rescale to \mbox{$\|A\|_F=\sqrt{k}$}. This changes the spectrum's shape at fixed norm. The resulting factor has condition number $\kappa$ and retains full column rank at $r\geq d$.

\textbf{Lifted AdamW (E34).} For fixed $A$ with full column rank, form $H=A(A^\top A)^{-1}$ and \mbox{$P=(A^\top A)^{-1}A^\top$} in float32. The update-space gradient is $\nabla_{\Delta W}\mathcal L=\nabla_B\mathcal L\,H$. We keep Adam's first and second moments on this $3d\times d$ matrix. With learning rate $\tau$, weight decay $\omega$ and the usual bias corrections, the step is
\[
D=-\tau\left(\frac{\hat m}{\sqrt{\hat v}+\epsilon}+\omega\Delta W\right),
\qquad B\leftarrow B+DP.
\]
Since $PA=I$, the change in $BA$ equals $D$ in exact arithmetic. The learning rate follows the main cosine schedule. Weight decay acts on $\Delta W$ here, while the dense baseline applies it to $W+\Delta W$.

\textbf{Lifted steps remove the condition-number effect on both backbones.} Every arm in Table~\ref{tab:frozen_backbones} uses the same three seeds, so each effect below is a paired difference. Under ordinary AdamW on $B$, raising $\kappa$ from $1$ to $5$ costs $0.065$ nats on DeiT-Tiny and $0.055$ on ViT-Tiny, and under lifted steps the difference falls below $10^{-3}$ nats on both. The lifted runs finish within $0.003$ nats of dense.

\begin{table}[ht]
\centering
\caption{Frozen-factor interventions at $r{=}2d$ on two backbones. Entries are mean test NLL and mean stable rank $\srank$ of the update over the same three seeds, using lr $5\times10^{-4}$ and $40$ epochs. All frozen-factor arms train $B$ and the head.}
\label{tab:frozen_backbones}
\begin{tabular}{lcrrrr}
\toprule
& & \multicolumn{2}{c}{DeiT-Tiny} & \multicolumn{2}{c}{ViT-Tiny} \\
\cmidrule(lr){3-4}\cmidrule(lr){5-6}
training coordinates & $\kappa(A)$ & test NLL & $\srank$ & test NLL & $\srank$ \\
\midrule
Dense qkv & --- & $1.9768$ & $29.2$ & $1.8097$ & $22.4$ \\
AdamW on $B$ & $1$ & $1.9716$ & $26.3$ & $1.8073$ & $19.4$ \\
AdamW on $B$ & $3$ & $2.0057$ & $22.4$ & $1.8385$ & $17.4$ \\
AdamW on $B$ & $5$ & $2.0367$ & $19.9$ & $1.8628$ & $16.2$ \\
Lifted AdamW & $1$ & $1.9789$ & $29.0$ & $1.8113$ & $22.2$ \\
Lifted AdamW & $5$ & $1.9789$ & $29.0$ & $1.8109$ & $22.2$ \\
\bottomrule
\end{tabular}
\end{table}

\subsection{Constraints on the trained factors}
\label{app:factorcons}

\textbf{Pinning $\kappa(A)$ by polar projection closes most of the full-rank gap (E37).} After each ordinary two-factor LoRA step, replace $A$ by $UV^\top$ from its thin SVD. This pins $\kappa(A)$ at $1$. With $B$ held fixed, the projection also changes $BA$ by about $5\%$ per step, comparable to the change from an optimizer step. At $r{=}d$ the projection closes $98\%$ of the gap to dense and roughly doubles the stable rank of the update, and at $2d$ the arm finishes below dense. At projection-weight parity it remains $0.040$ nats behind dense in every seed.

\textbf{Pinning $\kappa(A)$ while preserving $BA$ raises risk instead (E39).} We also absorb the stretch into $B$ while projecting $A$,
\[
A\leftarrow UV^\top,\qquad B\leftarrow B\bigl(U\Sigma U^\top+I-UU^\top\bigr).
\]
This preserves $BA$ to within $2\times10^{-6}$ relative error while setting $\kappa(A){=}1$. At $r{=}d$, NLL reaches $2.357$, or $0.380$ above dense. Both projections produce the same condition number but different risk. E39 also changes the parameter coordinates without transforming Adam's stored moments, so it does not isolate the effect of removing E37's function perturbation. The two results establish that condition number alone is insufficient to account for the improvement.

\textbf{Matching dense's stable rank does not match its risk (E29).} We add $\lambda$ times the mean over adapted layers of $\|\Delta W\|_2^2/\|\Delta W\|_F^2$, the inverse stable rank, to the training loss, with $\lambda\in\{0.03,0.3,1.0\}$ at $r\in\{64,192\}$ and three seeds. The penalty is invariant to the scale of $\Delta W$, so it raises stable rank without shrinking the update. At $r{=}d$ and $\lambda{=}1$ the penalty lifts stable rank from $11.2$ to $40.3$, above dense's $29.2$, while test NLL falls by about a tenth of the distance to the $r{=}1$ valley. At $\lambda{=}0.3$ the stable rank of the adapter nearly matches dense's, yet the adapter remains $0.070$ nats behind dense.
\FloatBarrier

\subsection{Training NLL against nominal rank}
\label{app:fitstatus}

\begin{figure}[ht]
\centering
\begin{minipage}[t]{0.36\linewidth}
\centering
\includegraphics[width=\linewidth]{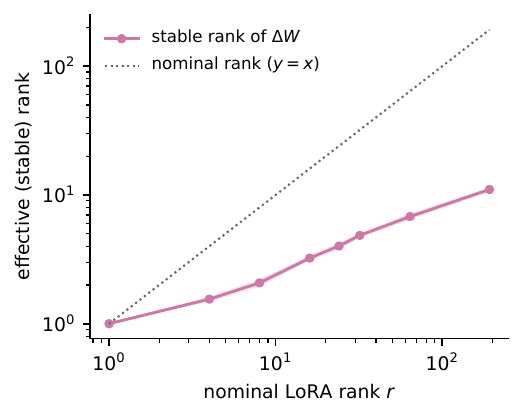}
\end{minipage}\hfill
\begin{minipage}[t]{0.60\linewidth}
\centering
\includegraphics[width=\linewidth]{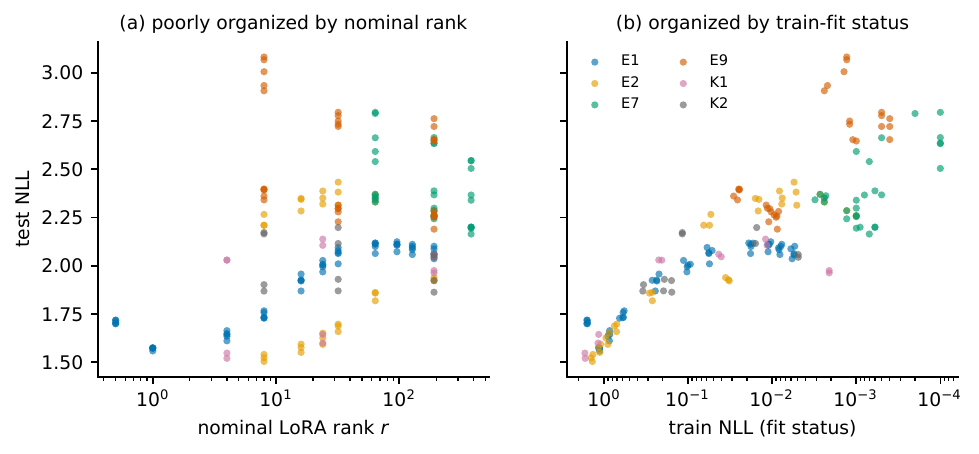}
\end{minipage}
\caption{\textbf{Left (effective complexity lags nominal rank).} Stable rank of the learned update vs.\ nominal rank ($20\%$ noise, $40$ ep); growth is sublinear everywhere and reaches only $11$ at $r{=}192$ ($y=x$ dotted). \textbf{Right (what organizes risk).} All $175$ noisy CIFAR-100 LoRA runs of the fixed-setting sweeps (E1/E2/E7/E9/K1/K2); test NLL against nominal rank (panel a, Spearman $\rho{=}0.40$) and against train NLL (panel b, $\rho{=}{-}0.86$, axis reversed, tie-corrected). Training NLL predicts test risk more closely than nominal rank does.}
\label{fig:unifying}
\end{figure}

Training NLL orders test risk more closely than nominal rank does in both pools of runs below, and in the fixed-setting pool also after the training budget is partialed out. The fixed-setting pool holds the $175$ noisy runs in which the backbone, prior and $n$ remain fixed while rank, epoch, learning rate and scaling vary (Figure~\ref{fig:unifying}, right). The broad pool holds all $486$ ordinary LoRA runs on DeiT-Tiny under $20\%$ synthetic noise, which also vary the prior, the data size, the placement and the optimizer settings. Train and test NLL are mechanically coupled before interpolation, so the second row partials the training budget out of the fixed-setting pool, which attenuates the ordering but leaves it intact.

\begin{center}
\footnotesize
\begin{tabular}{lcc}
\toprule
& \multicolumn{2}{c}{Spearman $\rho$ of test NLL with} \\
\cmidrule(lr){2-3}
pool of runs & nominal rank & train NLL \\
\midrule
fixed-setting ($175$ runs) & $0.40$ & $-0.86$ \\
fixed-setting, training budget partialed out & $0.31$ & $-0.55$ \\
broad ($486$ runs) & $0.37$ & $-0.43$ \\
\bottomrule
\end{tabular}
\end{center}

\textbf{Within most interpolated groups, higher rank mildly helps.} In six of the eight fully interpolated strata, grouped by block and schedule, the correlation with rank turns negative, which matches the post-peak descent of Section~\ref{sec:recovery}. The other two are the $40$- and $80$-epoch rank sweeps, whose interpolated ranks still include the climb to the peak. These are associations across runs that share seeds, subsets and schedules, not an identified effect.

\section{Breadth and robustness}
\label{app:breadth}

The cliff reappears on dense capacity axes, with human label noise and with bottleneck adapters, and risk also rises past the valley on a text task (Appendix~\ref{app:robust}). Per-rank model selection shrinks the cliff without removing it (Appendix~\ref{app:e24}), and corrupting the pretrained prior only deepens it (Appendix~\ref{app:prior}).

\subsection{Robustness panels}
\label{app:robust}

\begin{figure}[ht]
\centering
\includegraphics[width=0.57\linewidth]{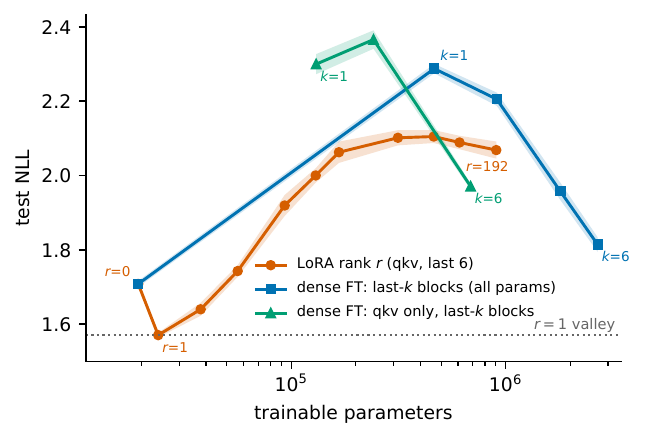}
\caption{\textbf{Dense axes recover more of the same cliff.} Test NLL vs.\ trainable parameters ($20\%$ noise, $40$ ep) along three capacity axes, namely LoRA rank (orange), dense fine-tuning of the last $k$ blocks (blue), and dense fine-tuning of only the qkv projections of the last $k$ blocks (green, module-matched to LoRA). All three climb a cliff once they interpolate; the dense axes then recover much of it, the low-rank axis almost none by $r{=}d$, and none reaches the $r{=}1$ valley (dotted).}
\label{fig:axes}
\end{figure}

\textbf{Dense axes recover more of the same cliff.} Figure~\ref{fig:axes} places LoRA rank beside two dense capacity axes on a common parameter scale (numbers in Table~\ref{tab:dense}). All three climb a cliff once they interpolate, and only the dense axes recover much of it.

\begin{figure}[ht]
\centering
\includegraphics[width=\linewidth]{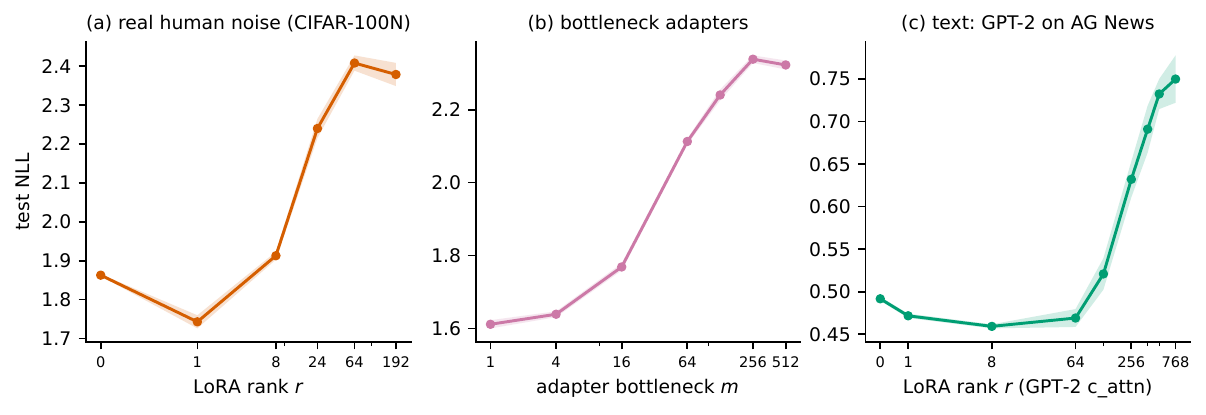}
\caption{\textbf{Breadth checks} (mean$\pm$std, $3$ seeds). (a)~Real human label noise (CIFAR-100N, ${\approx}40\%$); the cliff replicates with larger amplitude and reaches $\mathrm{Rec}=0.04$ by $d$ ($0.09$ by $2d$; \mbox{Table~\ref{tab:breadth}}). (b)~Bottleneck adapters; the same cliff along the adapter-width axis, peaking at $m{=}256$ and reaching only $\mathrm{Rec}=0.02$ by $m{=}512$. (c)~GPT-2 LoRA on noisy AG News; risk rises past the $r{=}8$ valley, and full rank, the only interpolating point, is also the largest capacity on the axis.}
\label{fig:breadth}
\end{figure}

\textbf{The rise persists across noise sources, adapter families and modalities.} We repeat the sweep with human label noise, bottleneck adapters and a GPT-2 text task (Figure~\ref{fig:breadth}).

\begin{figure}[ht]
\centering
\includegraphics[width=0.85\linewidth]{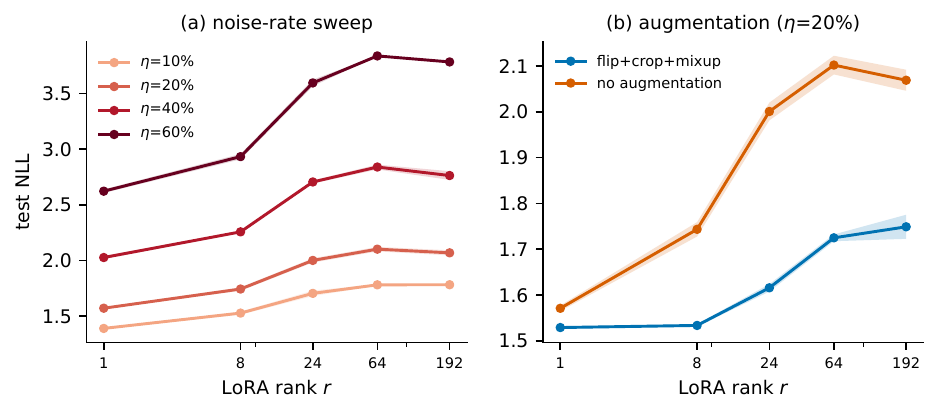}
\caption{\textbf{Noise rate and augmentation.} (a)~Test NLL vs.\ rank at $\eta \in \{10,20,40,60\}\%$; the valley remains at $r{=}1$ and the cliff amplitude grows monotonically with the noise rate. (b)~Augmentation (per-image flip, batch-shared pad-crop, mixup) lowers the curve and shrinks the cliff without removing it; the augmented curve shows no post-peak descent.}
\label{fig:noiseaug}
\end{figure}

\textbf{Noise rate and augmentation change only the depth of the cliff.} The valley remains at $r{=}1$ at every noise rate, and augmentation shrinks the cliff without removing it (Figure~\ref{fig:noiseaug}).

\begin{figure}[ht]
\centering
\includegraphics[width=\linewidth]{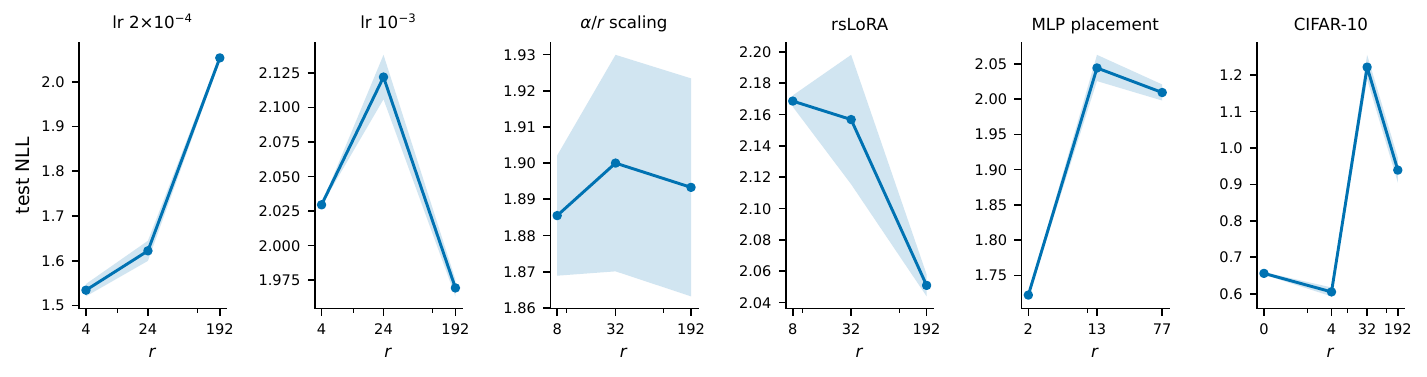}
\caption{\textbf{The cliff persists through design perturbations.} Test NLL vs.\ rank ($20\%$ noise unless noted) under learning rates $2\times10^{-4}$ and $10^{-3}$, standard $\alpha/r$ at $40$ ep (flat only because the high ranks have barely interpolated; Figure~\ref{fig:rankepoch}b trains them longer), rsLoRA, MLP placement at matched parameter budget (effective ranks $\{2,13,77\}$), and CIFAR-10.}
\label{fig:robust}
\end{figure}

\textbf{Design perturbations do not remove the cliff.} We vary the learning rate, the scaling rule, the adapter placement and the dataset, and the cliff persists under each change once the ranks interpolate (Figure~\ref{fig:robust}).

\subsection{Hyperparameter selection at each rank}
\label{app:e24}
Selecting the learning rate, weight decay and stopping epoch separately for each rank reduces the cliff from $0.53$ to $0.16$ nats but does not remove it. The tuned envelope still rises with rank, no configuration returns to the $r{=}1$ valley in any seed, and every selected checkpoint stops before interpolation.

\textbf{Selection.} For each rank we searched learning rate $\in \{1,2,5\}\times10^{-4} \cup \{10^{-3}\}$ and weight decay $\in \{0.01, 0.1\}$ and selected the configuration by validation NLL on a clean held-out split of $1{,}000$ training-pool images (test labels untouched), with the stopping epoch of each $80$-epoch run chosen per seed by the same split. Table~\ref{tab:e24} reports each rank's winning configuration re-run on three fresh seeds held out of the selection; the selection seed is excluded from the mean.

\textbf{No rank above one reaches the $r{=}1$ valley in any fresh seed.} In every fresh seed the best rank above one ($r{=}8$ in two seeds, $r{=}24$ in one) finishes $0.005$ to $0.021$ nats above $r{=}1$, and $r{=}384$ finishes about $0.16$ nats above it. Resampling whole seeds ($10$k draws) gives nearly the same ranges; with three seeds these describe the spread, and the per-seed differences remain the primary statistic.

\textbf{Every selected checkpoint stops before interpolation.} The train accuracy of the selected checkpoints rises with rank but remains below $99\%$ at every rank (Table~\ref{tab:e24}). The tuned envelope therefore compares sub-interpolation models throughout, and its test NLL still rises with rank. Because the selection covers LoRA ranks only and includes no dense arm, it establishes the shape of the tuned rank axis and not the gap to dense under selection; Appendix~\ref{app:envelope} gives each arm its own rate.

\begin{table}[ht]
\caption{\textbf{Per-rank envelope (E24).} Each rank's validation-selected learning rate and weight decay, with the stopping epoch, best validation NLL, and test NLL over three fresh seeds held out of the selection (epoch, train acc, and val NLL are seed means; test NLL is mean$\pm$std). Test NLL is read at each seed's own best-validation epoch. Train accuracy is that of the selected checkpoint, read at the nearest logged epoch (the train history is logged every $5$ epochs, so within $3$ epochs of the selected one); the final-epoch ($80$) train accuracy is $\geq 99\%$ for every rank except $r{=}1$ ($88\%$).}
\label{tab:e24}
\centering
\footnotesize
\begin{tabular}{rcccccc}
\toprule
rank & sel.\ lr & sel.\ wd & sel.\ epoch & train acc (sel., \%) & val NLL & test NLL (mean$\pm$std) \\
\midrule
$1$   & $5\times10^{-4}$ & $0.1$  & $12$ & $67$ & $1.544$ & $1.512 \pm 0.009$ \\
$8$   & $5\times10^{-4}$ & $0.01$ & $10$ & $70$ & $1.552$ & $1.531 \pm 0.008$ \\
$24$  & $10^{-3}$        & $0.1$  & $4$  & $75$ & $1.562$ & $1.539 \pm 0.014$ \\
$64$  & $10^{-3}$        & $0.1$  & $3$  & $86$ & $1.618$ & $1.597 \pm 0.004$ \\
$192$ & $5\times10^{-4}$ & $0.1$  & $3$  & $86$ & $1.682$ & $1.657 \pm 0.004$ \\
$384$ & $5\times10^{-4}$ & $0.1$  & $3$  & $94$ & $1.694$ & $1.672 \pm 0.008$ \\
\bottomrule
\end{tabular}
\end{table}
\FloatBarrier

\subsection{Prior-strength corruption}
\label{app:prior}
\begin{figure}[ht]
\centering
\includegraphics[width=0.55\linewidth]{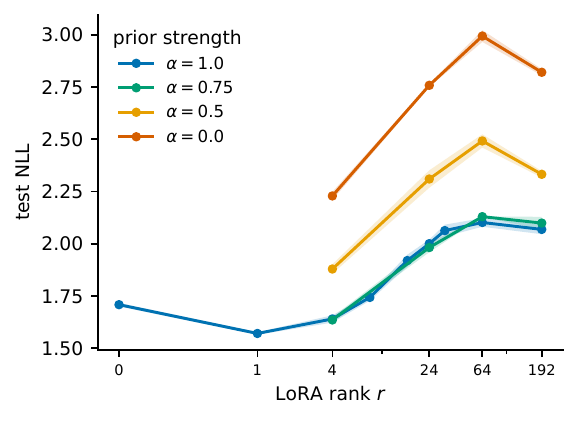}
\caption{\textbf{Weaker priors deepen the cliff.} Backbone corruption $W \leftarrow \alpha W_{\mathrm{pre}} + (1-\alpha)W_{\mathrm{rand}}$ on the adapted blocks ($20\%$ noise, $40$ ep, $3$ seeds; $5$ at $\alpha{=}1.0$) raises the peak and deepens the cliff monotonically. Peak-to-valley over the shared grid $r\in\{4,24,64,192\}$ is $0.46$, $0.49$, $0.61$, $0.76$ nats at $\alpha = 1.0$, $0.75$, $0.5$, $0.0$. No setting recovers its pre-peak minimum.}
\label{fig:prior}
\end{figure}

Mixing the adapted blocks' weights with random weights, $W \leftarrow \alpha W_{\mathrm{pre}} + (1-\alpha)W_{\mathrm{rand}}$, over $\alpha \in \{0.75, 0.5, 0.0\}$ at $r \in \{4,24,64,192\}$ and $3$ seeds, deepens the cliff monotonically relative to the uncorrupted main sweep ($\alpha{=}1.0$, $5$ seeds) and never repairs it (Figure~\ref{fig:prior}).

\textbf{A mild corruption deepens the cliff slightly and recovers no more of it.} The mildest corruption, $\alpha{=}0.75$, matters most, because it is the only corrupted setting where both ends of the grid remain interpretable. There the lowest rank still fits $75.5\%$ of the training set against $81.5\%$ at $\alpha{=}1.0$, so the left end is still a pre-interpolation valley rather than an underfitting artifact. Measured from $r{=}4$, the smallest rank of this grid, the cliff deepens from $0.46$ to $0.49$ nats and $\mathrm{Rec}$ falls from $0.07$ to $0.06$.

\textbf{Deeper corruptions change what the left end measures.} At $\alpha{=}0.5$ and $\alpha{=}0.0$ the lowest rank fits only $69\%$ and $57\%$ of the training set, so their left end measures underfitting a damaged feature space, not a valley. Consequently their $\mathrm{Rec}$ ($0.26$ and $0.23$) is not on the same footing as the axes in Table~\ref{tab:rec} and does not extend their ordering, which is claimed only for axes swept from a genuine pre-interpolation valley to a capacity limit. The finding that holds across all four settings and is used in the main text is that a weaker prior raises the peak and deepens the cliff without recovering the pre-peak minimum. Appendix~\ref{app:toy} isolates the role of prior quality in a warm-start linear model.

\FloatBarrier

\section{A minimal warm-start linear model}
\label{app:toy}

\begin{figure}[ht]
\centering
\includegraphics[width=\linewidth]{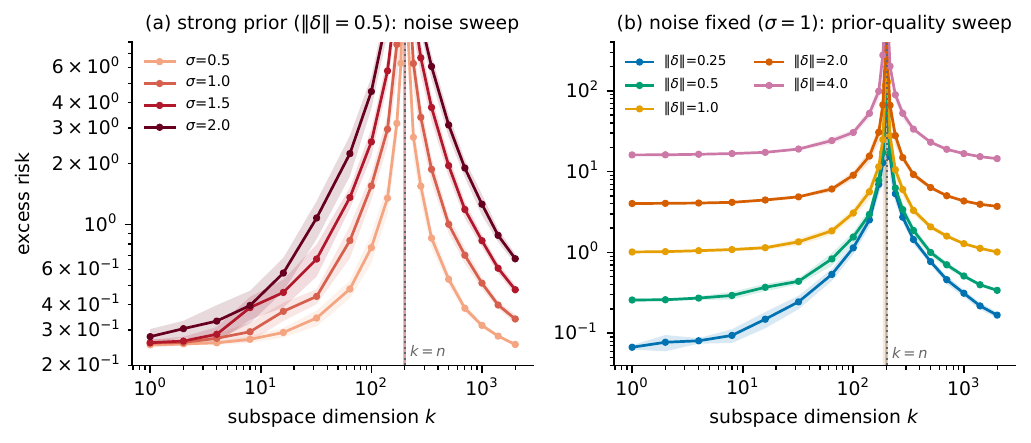}
\caption{\textbf{Warm-start linear model} ($p{=}2000$, $n{=}200$; mean$\pm$std over $16$ trials; the min-norm divergence at $k{=}n$ is clipped). (a)~With a strong prior ($\|\delta\|{=}0.5$), the plateau rises with the noise level while the valley remains anchored near $\|\delta\|^2$; the cliff amplitude grows with $\sigma^2$ and nearly vanishes at low noise, as in the experiments (Figures~\ref{fig:main}a,~\ref{fig:noiseaug}a). (b)~Degrading the prior at fixed noise morphs the cliff continuously into canonical double descent; for $\|\delta\| \lesssim \sigma$ the interpolating branch never reaches the valley, and for $\|\delta\| \gtrsim \sigma$ the second descent ends below it.}
\label{fig:toy}
\end{figure}

A warm-start linear model isolates how prior quality changes the height of the valley relative to the interpolating tail. Take overparameterized linear regression $y = \mathbf{x}^\top w^* + \varepsilon$ with $\mathbf{x} \sim \mathcal{N}(0, I_p)$, $\varepsilon \sim \mathcal{N}(0, \sigma^2)$, $n$ samples, and confine the trainable update to a random $k$-dimensional subspace around a warm start, so that $w = w_0 + S_k \alpha$ with $S_k \in \mathbb{R}^{p \times k}$ orthonormal and $w_0 = w^* + \delta$. Here $k$ plays the role of adapter rank, $\|\delta\|$ the prior error, and $\sigma$ the noise. Gradient descent from $\alpha = 0$ converges to the (minimum-norm) least-squares solution $\hat\alpha$, and the excess risk is $\|\delta + S_k \hat\alpha\|^2$. We simulate $p = 2000$, $n = 200$, $16$ trials per point (Figure~\ref{fig:toy}).

The expected excess risk follows from the ridgeless-regression calculation \citep{belkin2020two,hastie2022surprises}. For $k<n-1$, let $v_k=\|\delta\|^2(1-k/p)$ be the expected prior error outside the trainable subspace. This component also enters the regression residual as noise, giving
\[
R(k)=v_k+(\sigma^2+v_k)\frac{k}{n-k-1}.
\]
For $p>n$ and a nonzero prior error or noise variance, the expression increases with $k$, placing the pre-interpolation valley at the smallest subspace dimension. At full capacity, assuming $p>n+1$,
\[
R(p)=\|\delta\|^2\left(1-\frac np\right)+\sigma^2\frac{n}{p-n-1}.
\]
The tail rises with noise and crosses the warm-start risk $\|\delta\|^2$ when $\|\delta\|^2=\sigma^2p/(p-n-1)$. At $\sigma=1$, this gives $\|\delta\|\approx1.05$, consistent with the simulated curves. A strong prior leaves the tail above the valley; a sufficiently weak prior, with $\|\delta\|^2>\sigma^2p/(p-n-1)$, allows the second descent to finish below it.

Optimization here reaches the minimum-norm solution in the chosen subspace. The model separates the effect of prior quality from the factor dynamics studied in Section~\ref{sec:nominal}, and connects the observed curve shapes with subspace-constrained risk trade-offs \citep{zindari2026lora}.

\section{Full numeric tables}
\label{app:tables}

\begin{table}[ht]
\centering
\caption{Main sweep (E1), test NLL and accuracy vs.\ rank; mean$\pm$std over $5$ seeds, $40$ epochs. The rows $2d$--$4d$ are the noisy-label tail runs of Table~\ref{tab:recovery}, over $3$, $8$ and $3$ seeds. E1's $r{=}96$ and $128$ runs appear with the densified window in Table~\ref{tab:window}.}
\label{tab:e1}
\small
\begin{tabular}{lcccccc}
\toprule
& \multicolumn{3}{c}{$20\%$ noise} & \multicolumn{3}{c}{clean} \\
\cmidrule(lr){2-4}\cmidrule(lr){5-7}
$r$ & test NLL & test acc (\%) & train acc (\%) & test NLL & test acc (\%) & train acc (\%) \\
\midrule
0   & $1.709{\scriptstyle\pm.008}$ & $57.1$ & $66.5$ & $1.382{\scriptstyle\pm.008}$ & $61.2$ & $85.2$ \\
1   & $1.571{\scriptstyle\pm.006}$ & $60.4$ & $74.6$ & $1.186{\scriptstyle\pm.014}$ & $66.4$ & $94.8$ \\
4   & $1.640{\scriptstyle\pm.017}$ & $58.7$ & $81.5$ & $1.229{\scriptstyle\pm.014}$ & $65.8$ & $98.3$ \\
8   & $1.743{\scriptstyle\pm.016}$ & $57.1$ & $88.7$ & $1.273{\scriptstyle\pm.017}$ & $65.6$ & $99.7$ \\
16  & $1.920{\scriptstyle\pm.028}$ & $54.6$ & $98.0$ & $1.363{\scriptstyle\pm.019}$ & $65.3$ & $100$ \\
24  & $2.001{\scriptstyle\pm.019}$ & $53.7$ & $99.9$ & $1.410{\scriptstyle\pm.027}$ & $65.1$ & $100$ \\
32  & $2.063{\scriptstyle\pm.029}$ & $53.5$ & $100$ & $1.455{\scriptstyle\pm.029}$ & $65.0$ & $100$ \\
64  & $2.102{\scriptstyle\pm.020}$ & $53.9$ & $100$ & $1.502{\scriptstyle\pm.017}$ & $65.0$ & $100$ \\
192 & $2.069{\scriptstyle\pm.023}$ & $55.4$ & $100$ & $1.519{\scriptstyle\pm.014}$ & $65.7$ & $100$ \\
\midrule
$2d$ & $2.019{\scriptstyle\pm.014}$ & $56.0$ & $100$ & --- & --- & --- \\
$3d$ & $1.983{\scriptstyle\pm.032}$ & $56.8$ & $100$ & --- & --- & --- \\
$4d$ & $1.953{\scriptstyle\pm.012}$ & $57.5$ & $100$ & --- & --- & --- \\
\bottomrule
\end{tabular}
\end{table}

\begin{table}[ht]
\centering
\caption{Dense capacity axes ($20\%$ noise, $40$ epochs unless noted; mean$\pm$std over $5$ seeds, $2$ for the $80$-epoch rows). E11 unfreezes the last $k$ blocks entirely; E13 unfreezes only their qkv projections (module-matched to LoRA). E13's $k{=}6$ arm is a separate run, on five of the same seeds, of the configuration that Table~\ref{tab:recovery} reports over eight seeds ($1.979$) and Section~\ref{sec:nominal} over the three shared with the intervention arms ($1.977$).}
\label{tab:dense}
\small
\begin{tabular}{lccc}
\toprule
configuration & trainable params & test NLL & train acc (\%) \\
\midrule
head only ($k=0$)        & $19{,}300$    & $1.709{\scriptstyle\pm.008}$ & $66.5$ \\
E11 dense blocks $k=1$   & $464{,}164$   & $2.288{\scriptstyle\pm.013}$ & $100$ \\
E11 dense blocks $k=2$   & $909{,}028$   & $2.206{\scriptstyle\pm.018}$ & $100$ \\
E11 dense blocks $k=4$   & $1{,}798{,}756$ & $1.958{\scriptstyle\pm.023}$ & $100$ \\
E11 dense blocks $k=6$   & $2{,}688{,}484$ & $1.814{\scriptstyle\pm.020}$ & $100$ \\
E11 dense blocks $k=6$, $80$ ep & $2{,}688{,}484$ & $1.961{\scriptstyle\pm.006}$ & $100$ \\
\midrule
E13 dense qkv $k=1$      & $130{,}468$   & $2.300{\scriptstyle\pm.027}$ & $97.7$ \\
E13 dense qkv $k=2$      & $241{,}636$   & $2.366{\scriptstyle\pm.025}$ & $100$ \\
E13 dense qkv $k=6$      & $686{,}308$   & $1.971{\scriptstyle\pm.008}$ & $100$ \\
E13 dense qkv $k=6$, $80$ ep & $686{,}308$ & $2.140{\scriptstyle\pm.005}$ & $100$ \\
\midrule
LoRA $r=1$ (valley)      & $23{,}908$    & $1.571{\scriptstyle\pm.006}$ & $74.6$ \\
LoRA $r=192$             & $904{,}036$   & $2.069{\scriptstyle\pm.023}$ & $100$ \\
\bottomrule
\end{tabular}
\end{table}

\begin{table}[ht]
\centering
\caption{DeiT-Small rank sweep (E12), $20\%$ noise, $40$ epochs, mean$\pm$std over $5$ seeds ($3$ at $r{=}768$). The sweep runs to $2d$; the axis is back at its valley by $d{=}384$, this model's expressivity limit ($\mathrm{Rec}=0.99$), and below it at $2d$ in every seed ($\mathrm{Rec}=1.25$).}
\label{tab:e12}
\small
\begin{tabular}{lcccccc}
\toprule
$r$ & 1 & 24 & 64 & 192 & 384 & 768 \\
\midrule
test NLL & $1.396{\scriptstyle\pm.011}$ & $1.529{\scriptstyle\pm.038}$ & $1.522{\scriptstyle\pm.040}$ & $1.450{\scriptstyle\pm.038}$ & $1.397{\scriptstyle\pm.040}$ & $1.363{\scriptstyle\pm.017}$ \\
train acc (\%) & $91.7$ & $100$ & $100$ & $100$ & $100$ & $100$ \\
\bottomrule
\end{tabular}
\end{table}

\begin{table}[ht]
\centering
\caption{High-rank/long-time tail (E7, fixed scale; E9, $\alpha/r$ scale), $20\%$ noise, test NLL, mean$\pm$std over $5$ seeds.}
\label{tab:tail}
\small
\begin{tabular}{lccc@{\qquad}ccc}
\toprule
& \multicolumn{3}{c}{E7 (fixed)} & \multicolumn{3}{c}{E9 ($\alpha/r$)} \\
\cmidrule(lr){2-4}\cmidrule(lr){5-7}
epochs & $r{=}64$ & $r{=}192$ & $r{=}384$ & $r{=}8$ & $r{=}32$ & $r{=}192$ \\
\midrule
80  & $2.351{\scriptstyle\pm.014}$ & $2.269{\scriptstyle\pm.020}$ & $2.192{\scriptstyle\pm.014}$ & $2.378{\scriptstyle\pm.023}$ & $2.283{\scriptstyle\pm.030}$ & $2.249{\scriptstyle\pm.031}$ \\
160 & $2.676{\scriptstyle\pm.103}$ & $2.538{\scriptstyle\pm.131}$ & $2.460{\scriptstyle\pm.089}$ & $2.999{\scriptstyle\pm.070}$ & $2.756{\scriptstyle\pm.027}$ & $2.688{\scriptstyle\pm.046}$ \\
\bottomrule
\end{tabular}
\end{table}

\begin{table}[ht]
\centering
\caption{Calibration and uncertainty at key ranks ($20\%$ noise unless noted), mean$\pm$std over $5$ seeds. Temperature fit on a held-out clean train split; test labels untouched. ECE is expected calibration error and AURC the area under the risk--coverage curve.}
\label{tab:calib}
\small
\begin{tabular}{lcccccc}
\toprule
$r$ & NLL raw & NLL calibrated & $T$ & ECE & Brier & AURC \\
\midrule
1 (valley) & $1.571{\scriptstyle\pm.006}$ & $1.535{\scriptstyle\pm.007}$ & $0.845{\scriptstyle\pm.010}$ & $.080{\scriptstyle\pm.004}$ & $.542{\scriptstyle\pm.004}$ & $.186{\scriptstyle\pm.004}$ \\
64 (peak)  & $2.102{\scriptstyle\pm.020}$ & $1.979{\scriptstyle\pm.018}$ & $1.346{\scriptstyle\pm.014}$ & $.130{\scriptstyle\pm.003}$ & $.628{\scriptstyle\pm.004}$ & $.230{\scriptstyle\pm.005}$ \\
192        & $2.069{\scriptstyle\pm.023}$ & $1.932{\scriptstyle\pm.019}$ & $1.373{\scriptstyle\pm.012}$ & $.137{\scriptstyle\pm.004}$ & $.616{\scriptstyle\pm.006}$ & $.214{\scriptstyle\pm.002}$ \\
\midrule
clean, 1   & $1.186{\scriptstyle\pm.014}$ & $1.177{\scriptstyle\pm.013}$ & $1.109{\scriptstyle\pm.005}$ & --- & --- & --- \\
clean, 192 & $1.519{\scriptstyle\pm.014}$ & $1.344{\scriptstyle\pm.012}$ & $1.471{\scriptstyle\pm.013}$ & --- & --- & --- \\
\bottomrule
\end{tabular}
\end{table}

\begin{table}[ht]
\centering
\caption{Breadth checks, per configuration (mean$\pm$std, $3$ seeds). The top block is CIFAR-100N real human noise (E14), the middle block is bottleneck adapters (E16) under $20\%$ synthetic noise, and the bottom block is GPT-2 LoRA on AG News with $20\%$ noise (E15).}
\label{tab:breadth}
\small
\begin{tabular}{lccc}
\toprule
configuration & test NLL & test acc (\%) & train acc (\%) \\
\midrule
E14 $r=0$    & $1.863{\scriptstyle\pm.003}$ & $50.6$ & $65.7$ \\
E14 $r=1$ (valley) & $1.743{\scriptstyle\pm.017}$ & $52.9$ & $72.7$ \\
E14 $r=8$    & $1.913{\scriptstyle\pm.010}$ & $50.5$ & $89.1$ \\
E14 $r=24$   & $2.240{\scriptstyle\pm.025}$ & $48.2$ & $99.9$ \\
E14 $r=64$ (peak) & $2.408{\scriptstyle\pm.020}$ & $48.0$ & $100$ \\
E14 $r=192$  & $2.378{\scriptstyle\pm.030}$ & $49.5$ & $100$ \\
E14 $r=384$  & $2.350{\scriptstyle\pm.027}$ & $50.2$ & $100$ \\
\midrule
E16 $m=1$ (valley) & $1.611{\scriptstyle\pm.011}$ & $59.2$ & $74.1$ \\
E16 $m=4$    & $1.639{\scriptstyle\pm.008}$ & $58.9$ & $77.9$ \\
E16 $m=16$   & $1.769{\scriptstyle\pm.010}$ & $56.7$ & $88.3$ \\
E16 $m=64$   & $2.113{\scriptstyle\pm.008}$ & $52.8$ & $99.8$ \\
E16 $m=128$  & $2.241{\scriptstyle\pm.015}$ & $52.2$ & $100$ \\
E16 $m=256$ (peak) & $2.339{\scriptstyle\pm.010}$ & $52.4$ & $100$ \\
E16 $m=512$  & $2.323{\scriptstyle\pm.012}$ & $52.7$ & $100$ \\
\midrule
E15 $r=0$    & $0.492{\scriptstyle\pm.002}$ & $88.8$ & $73.2$ \\
E15 $r=1$    & $0.472{\scriptstyle\pm.002}$ & $89.6$ & $74.2$ \\
E15 $r=8$ (valley) & $0.459{\scriptstyle\pm.002}$ & $89.4$ & $75.3$ \\
E15 $r=64$   & $0.469{\scriptstyle\pm.011}$ & $86.8$ & $81.3$ \\
E15 $r=128$  & $0.521{\scriptstyle\pm.019}$ & $83.9$ & $87.4$ \\
E15 $r=256$  & $0.632{\scriptstyle\pm.021}$ & $80.9$ & $95.0$ \\
E15 $r=384$  & $0.691{\scriptstyle\pm.028}$ & $80.2$ & $97.7$ \\
E15 $r=512$  & $0.732{\scriptstyle\pm.018}$ & $79.9$ & $98.8$ \\
E15 $r=768$  & $0.750{\scriptstyle\pm.028}$ & $80.5$ & $99.4$ \\
\bottomrule
\end{tabular}
\end{table}

\begin{table}[ht]
\centering
\caption{Backbone external validity (E26), per configuration (mean$\pm$std, $3$ seeds), CIFAR-100 $5$k, $20\%$ noise, $40$ epochs. ViT-Tiny uses the AugReg ImageNet-21k$\to$ImageNet-1k checkpoint \citep{steiner2022augreg} with its native $0.5$ normalization; on Swin-Tiny, LoRA adapts the last six blocks (stage dims $384/768$), so every adapted block reaches full rank by $r{=}768$. On Swin every $r \geq 1$ interpolates and accuracy remains above its $r{=}1$ level while NLL climbs.}
\label{tab:e26}
\small
\begin{tabular}{lccc}
\toprule
configuration & test NLL & test acc (\%) & train acc (\%) \\
\midrule
ViT-Tiny $r=0$    & $1.575{\scriptstyle\pm.002}$ & $60.0$ & $77.9$ \\
ViT-Tiny $r=1$ (valley) & $1.537{\scriptstyle\pm.009}$ & $61.2$ & $83.9$ \\
ViT-Tiny $r=24$   & $1.899{\scriptstyle\pm.017}$ & $57.4$ & $100$ \\
ViT-Tiny $r=96$ (peak) & $1.902{\scriptstyle\pm.039}$ & $58.7$ & $100$ \\
ViT-Tiny $r=192$  & $1.871{\scriptstyle\pm.018}$ & $59.4$ & $100$ \\
\midrule
Swin-Tiny $r=0$   & $1.483{\scriptstyle\pm.010}$ & $63.9$ & $77.5$ \\
Swin-Tiny $r=1$ (valley) & $1.353{\scriptstyle\pm.014}$ & $65.8$ & $99.2$ \\
Swin-Tiny $r=96$  & $1.378{\scriptstyle\pm.027}$ & $68.3$ & $100$ \\
Swin-Tiny $r=384$ & $1.512{\scriptstyle\pm.019}$ & $67.1$ & $100$ \\
Swin-Tiny $r=768$ & $1.602{\scriptstyle\pm.024}$ & $66.2$ & $100$ \\
\bottomrule
\end{tabular}
\end{table}

\begin{table}[ht]
\centering
\caption{Data-size sweep (E10), $20\%$ noise, $40$ epochs, test NLL. The $n{=}5{,}000$ row is the $5$-seed main sweep; the other two rows average $2$ seeds.}
\label{tab:e10}
\small
\begin{tabular}{lccccc}
\toprule
$n$ & $r{=}1$ & $r{=}8$ & $r{=}24$ & $r{=}64$ & $r{=}192$ \\
\midrule
$2{,}500$  & $1.775$ & $1.880$ & $2.076$ & $2.185$ & $2.172$ \\
$5{,}000$  & $1.571$ & $1.743$ & $2.001$ & $2.102$ & $2.069$ \\
$10{,}000$ & $1.433$ & $1.645$ & $2.085$ & $2.168$ & $2.086$ \\
\bottomrule
\end{tabular}
\end{table}
\FloatBarrier

\end{document}